\documentclass{article}

\PassOptionsToPackage{numbers, compress}{natbib}
 \usepackage[preprint]{neurips_2026}
\usepackage{wrapfig}
\usepackage[utf8]{inputenc} % allow utf-8 input
\usepackage[T1]{fontenc}    % use 8-bit T1 fonts
\usepackage{hyperref}       % hyperlinks
\usepackage{url}            % simple URL typesetting
\usepackage{booktabs}       % professional-quality tables
\usepackage{amsfonts}       % blackboard math symbols
\usepackage{nicefrac}       % compact symbols for 1/2, etc.
\usepackage{microtype}      % microtypography
\usepackage{xcolor}         % colors
\usepackage{multirow} 
\usepackage{amsmath}
\usepackage[table]{xcolor}
\usepackage{pifont}
\usepackage{amssymb}
\usepackage{graphicx}   % resizebox
\newtheorem{definition}{Definition}
\usepackage{xspace}
\usepackage{subcaption}

\usepackage[ruled,vlined,linesnumbered]{algorithm2e}
\SetKwInput{KwIn}{Input}
\SetKwInput{KwOut}{Output}
\newcommand{\ourmethod}{\textsc{{CIE-Scorer}}\xspace}

\title{Detecting Unfaithful Chain-of-Thought via Circuit-Guided Internal-External Discrepancy} %internal is adj, need to adjust

\author{%
  Xu Shen\textsuperscript{1}, Zhen Tan\textsuperscript{2}, Song Wang\textsuperscript{3}, Pingjun Hong\textsuperscript{4}, Rui Miao\textsuperscript{1}, Xin Wang \textsuperscript{1},Tianlong Chen\textsuperscript{5}\\
  \textsuperscript{1}Jilin University
  \textsuperscript{2}University of Central Florida  
  \textsuperscript{3}Arizona State University \\
  \textsuperscript{4}University of Vienna 
  \textsuperscript{5}University of North Carolina at Chapel Hill \\
}

\begin{document}

\maketitle

\begin{abstract}
Chain-of-thought (CoT) reasoning improves the problem-solving ability of large language models (LLMs), but generated reasoning traces may not faithfully reflect the model's actual decision process. Existing CoT unfaithfulness detectors mainly rely on external signals from generated rationales, such as textual plausibility or answer consistency, while overlooking evidence from the model's internal computation. 
%\sw{I feel like this sentence is too vague. What is generated rationales? why is it in contrast to internal computation?} 
Although recent circuit tracing methods provide a way to obtain model-internal evidence by tracing how information flows through model components during reasoning, 
%\sw{by looking at xxx or extracting xxx}
constructing full reasoning circuits for long CoTs is costly and difficult to scale. 
%\sw{what is circuit? how to construct it? why is it not scalable? we should explain these with short words} 
To address these challenges, 
%\sw{what challenge are you going to address? if it is scalability, then we need to mention in the following explicitly that we propose an efficient design and explain why it is efficient} 
we propose \textbf{C}ircuit-guided \textbf{I}nternal-\textbf{E}xternal Discrepancy Scorer (\ourmethod), a framework for instance-level CoT unfaithfulness detection. 
%Specifically, \ourmethod first selects informative tokens within each reasoning sentence and traces circuits only for these tokens, yielding compact sentence-level circuit graphs. These graphs are then compressed into internal reasoning embeddings. In parallel, \ourmethod extracts external sentence representations from LLM hidden states and constructs an external reasoning graph under the same graph-building rule. Finally, it computes the fused Gromov--Wasserstein distance between the internal and external reasoning graphs, using their discrepancy as the unfaithfulness score. 
The key idea is that faithful reasoning traces should align with the model's computational process, whereas unfaithful traces may diverge from it. \ourmethod efficiently traces compact sentence-level circuits from informative reasoning tokens, constructs internal and external reasoning graphs, and measures their discrepancy using Fused Gromov--Wasserstein distance. 
Experiments on four datasets from FaithCoT-Bench show that \ourmethod achieves state-of-the-art performance while reducing the cost of circuit construction, demonstrating the effectiveness of combining mechanistic interpretability signals with external reasoning traces for CoT unfaithfulness detection.The code is provide in 
\href{https://github.com/se7esx/CIE-Scorer}{https://github.com/se7esx/CIE-Scorer}.
\end{abstract}

\section{Introduction}
Chain-of-thought (CoT) reasoning has become a central mechanism for improving the reasoning ability of large language models (LLMs), by enabling them to produce intermediate textual steps before committing to a final answer~\citep{cot,zero_cot}. Such reasoning traces have led to substantial performance gains in mathematics, programming, logic, and commonsense reasoning~\citep{math_intro,code_intro,logic_intro,common_intro}, and are treated as an intuitive window into the model's reasoning process~\citep{tot,got,long_cot}. However, recent studies have shown that this \textit{apparent transparency} can be misleading: although CoT traces resemble step-by-step reasoning in natural language, they may not correspond to the internal mechanisms that underlie the model's predictions~\citep{feng2023towards,zhao2025chain,not_explainability,matton2025walk}. 
%\sw{reveal what computation? CoT are texts, how do they "reveal" computation? If you want to say CoT is different from model internal reasoning, then we should make it explicit} 
Existing works have examined surface-level reliability concerns, including hallucination, factual correctness, and rationale quality~\citep{llm_safety,safty_align,llm_trustworthiness,park2025steer}. These concerns are related to the notion of \textbf{\textit{faithfulness}}, which asks whether an explanation reflects the model's internal decision process~\citep{shen2025faithcot,lyu2023faithful}. An \textit{\textbf{unfaithful}} explanation may be coherent and persuasive while providing misleading accounts of model behavior, limiting its usefulness for understanding or auditing model behavior and potentially leading to safety concerns~\citep{turpin2023language,Agarwal2024Faithfulness}. 
%\sw{I dont see why faithfulness is fundamental requirement here. we should emphasize its importance with examples or applications, something like unfaithfulness will lead to what consequences} 
Therefore, determining whether CoT reasoning is faithful is essential for evaluating its role as an explanation of LLM behavior.

Recent studies examine CoT faithfulness from two main perspectives. One line of work evaluates the causal strength between CoT traces and final predictions through counterfactual interventions
%\sw{explain what is counterfactual intervention} 
, where parts of the CoT are perturbed to examine their effect on the final prediction~\citep{lanham2023measuring,turpin2023language,yang2025well,xiong2025measuring}. While effective for revealing whether CoT functions as a faithful explanation, such analysis does not directly determine the faithfulness of a specific generated trace. A second line evaluates reasoning traces through LLM-as-Judge or process reward models, treating erroneous reasoning with correct answers or plausible reasoning with incorrect answers as indicators of unfaithfulness~\citep{wild_faithful,step_judge}.However, {reasoning correctness} and {faithfulness} are not equivalent: a reasoning trace may arrive at the correct answer while failing to reflect the model's actual decision process.
%\sw{so what is the difference between reasoning correctness and faithfulness?} 
Since faithfulness needs to be assessed for individual reasoning traces, \citet{shen2025faithcot} introduces the \textsc{FINE-CoT} dataset for training and evaluating CoT unfaithfulness detectors, which we use as the primary benchmark in our study. %\sw{why do we mention this dataset here} 
Taken together, current detection methods for CoT unfaithfulness still rely largely on the {external content} of generated reasoning traces, leaving the model's internal computation underexplored. This raises \textit{\textbf{Challenge 1:}} \textit{how can we detect CoT unfaithfulness using evidence from the model's internal decision process rather than only surface-level reasoning quality?}

White-box interpretability offers a way to move beyond \textbf{surface-level} reasoning quality by examining the internal mechanisms~\citep{zhao2024explainability,du2019techniques}. In particular, circuit tracing methods analyze how information flows through internal model components during reasoning, allowing us to identify which internal features and interactions contribute to the prediction~\citep{gaoscaling,hannahave,ameisen2025circuit}.
%\sw{what feature representations? what is causal subgraph? what is circuit analysis?}
%\sw{what is transcoder? what is internal features?} 
These internal traces offer a promising source of evidence for detecting CoT unfaithfulness. However, existing mechanistic interpretability pipelines remain computationally demanding and difficult to scale, especially for long reasoning traces. Constructing attribution graphs across many reasoning steps requires tracing large numbers of internal interactions, making full circuit construction costly.
%\sw{why they are unscalable?} 
Although recent work uses step-level circuit topology to train classifiers for reasoning correctness~\citep{crv}, it is limited to short reasoning traces. 
%\sw{why it is costly} 
This gives rise to \textit{\textbf{Challenge 2:}} \textit{how can we exploit internal interpretability signals for CoT unfaithfulness detection while avoiding the prohibitive cost of full circuit construction?}

Based on the insights above, we propose \textbf{C}ircuit-guided \textbf{I}nternal-\textbf{E}xternal Discrepancy Scorer (\ourmethod) for CoT unfaithfulness detection. 
%\sw{until now, you have not defined what is "unfaithfulness", what are we detecting?} 
To address \textit{\textbf{Challenge 1}}, 
%\sw{why this can solve challenge 1?} 
\ourmethod leverages circuit-level evidence to characterize the model's internal computational trajectory, allowing unfaithfulness to be detected from mismatch between internal computation and externally reasoning traces. To address \textit{\textbf{Challenge 2}}, we select informative tokens in each reasoning sentence before circuit tracing, enabling analysis without constructing full circuits over CoTs. The resulting sentence-level computational graphs form a circuit sequence, which is compressed into internal reasoning embeddings. We then compare this internal trajectory with the external reasoning trajectory by constructing two sentence graphs under a shared graph-building rule: an internal graph from the compressed circuit embeddings and an external graph from hidden representations of the displayed reasoning sentences. The key intuition is that faithful reasoning traces should align with the model's internal computational process, whereas unfaithful traces may diverge from it. After aligning the two latent spaces with a lightweight adaptor, we compute the {Fused Gromov--Wasserstein (FGW) distance} between the graphs and use it as the unfaithfulness score. 
%\sw{why we want to do this combine? what is the benefit?} 
To the best of our knowledge, \ourmethod is the first approach to explicitly combine the model's internal computational process with its externally displayed reasoning trace. Experiments on four datasets from FaithCoT-Bench demonstrate that \ourmethod achieves state-of-the-art performance. Our main contributions are summarized as follows:
\begin{itemize}
    \item We formulate CoT unfaithfulness detection from a \textbf{circuit-guided internal-external discrepancy} perspective, to our knowledge, the first to integrate model-internal computation with external reasoning traces for instance-level faithfulness analysis.

    \item We propose \ourmethod{}, which performs token-selected circuit tracing, compresses sentence-level circuit graphs into internal reasoning embeddings, and compares them with external reasoning representations using Fused Gromov--Wasserstein distance.

    \item Experiments on four FaithCoT-Bench datasets show that \ourmethod{} achieves state-of-the-art performance against representative baselines, validating the effectiveness of efficient circuit-guided detection.
\end{itemize}

\section{Preliminaries}
\subsection{CoT Unfaithfulness Detection}
CoT reasoning generates intermediate reasoning steps before producing a final answer. Given a query $q$ and a prompting instruction $p$, a large language model $\mathcal{M}$ produces a reasoning trajectory $\mathcal{C}=(r_1,r_2,\ldots,r_T)$, where $r_t$ denotes the $t$-th reasoning step and $T$ is the trajectory length, followed by a final answer $a$. Although $\mathcal{C}$ is often treated as an explanation for $a$, it may not faithfully reflect the model's actual decision process. Prior studies have mainly examined CoT unfaithfulness at the population level, using counterfactual interventions to test whether CoT explanations are causally linked to final answers.  In contrast, \citet{shen2025faithcot} formulates CoT faithfulness as an instance-level detection problem, where each query--CoT instance is assigned a binary label based on observable signals of unfaithfulness.
\begin{definition}[Instance-level CoT Unfaithfulness Detection]
Given a query $q$ and a generated CoT trajectory $\mathcal{C}=(r_1,r_2,\ldots,r_T)$ produced by a large language model $\mathcal{M}$, CoT unfaithfulness detection aims to decide whether $\mathcal{C}$ faithfully reflects the model's internal reasoning process, denoted as $\mathcal{R}_{\mathcal{M}}(q)$. Formally, this task is defined as a binary classification function:
\begin{equation}
    f: (q,\mathcal{C}) \mapsto \{0,1\},
\end{equation}
where $f(q,\mathcal{C})=1$ indicates that $\mathcal{C}$ is unfaithful to $\mathcal{R}_{\mathcal{M}}(q)$, and $f(q,\mathcal{C})=0$ indicates that $\mathcal{C}$ is faithful. Different detection methods instantiate $f$ using different observable evidence.
\end{definition}

\subsection{Circuit Tracing}
We adopt circuit tracing~\citep{ameisen2025circuit} to obtain an interpretable approximation of the model's internal computation. In mechanistic interpretability, a circuit is typically viewed as a subgraph of model components and interactions that supports a specific computation. Circuit tracing constructs such graphs on a transcoder-based replacement model, where dense MLP activations are decomposed into sparse overcomplete features, enabling feature-level attribution. Given a prompt, it produces a prompt-specific attribution graph
$\mathcal{G}_{\mathrm{attr}}=(\mathcal{V},\mathcal{E})$,
where $\mathcal{V}$ contains input nodes, feature nodes, and output nodes, and $\mathcal{E}$ contains directed weighted edges representing causal contributions between nodes.

Here, input nodes correspond to prompt tokens, output nodes correspond to candidate next-token logits, and feature nodes correspond to sparse transcoder features activated at specific layers and token positions. Formally, the set of active feature nodes is
\begin{equation}
    \mathcal{V}_{f}
    =
    \left\{
    v=(\ell,t,k)
    \mid
    f_{\ell,t,k} > 0
    \right\},
\end{equation}
where $\ell$, $t$, and $k$ denote the layer index, token position, and feature index, respectively. For a directed edge from a source feature $s$ to a target feature or output node $t$, the attribution weight is computed as
\begin{equation}
    A_{s\rightarrow t}
    =
    a_s w_{s\rightarrow t},
\end{equation}
where $a_s$ is the activation of the source feature and $w_{s\rightarrow t}$ denotes its effective linear influence on the target under the local replacement model. Thus, the resulting graph decomposes a next-token prediction into sparse feature activations and their directed contributions. In this work, we use circuit tracing as a graph-construction operator, denoted by $\mathrm{Trace}(\cdot)$, and later apply it to selected reasoning tokens to obtain compact sentence-level circuit graphs.

\section{Method}
\label{sec:method}
\begin{figure}
    \centering
    \includegraphics[width=\linewidth]{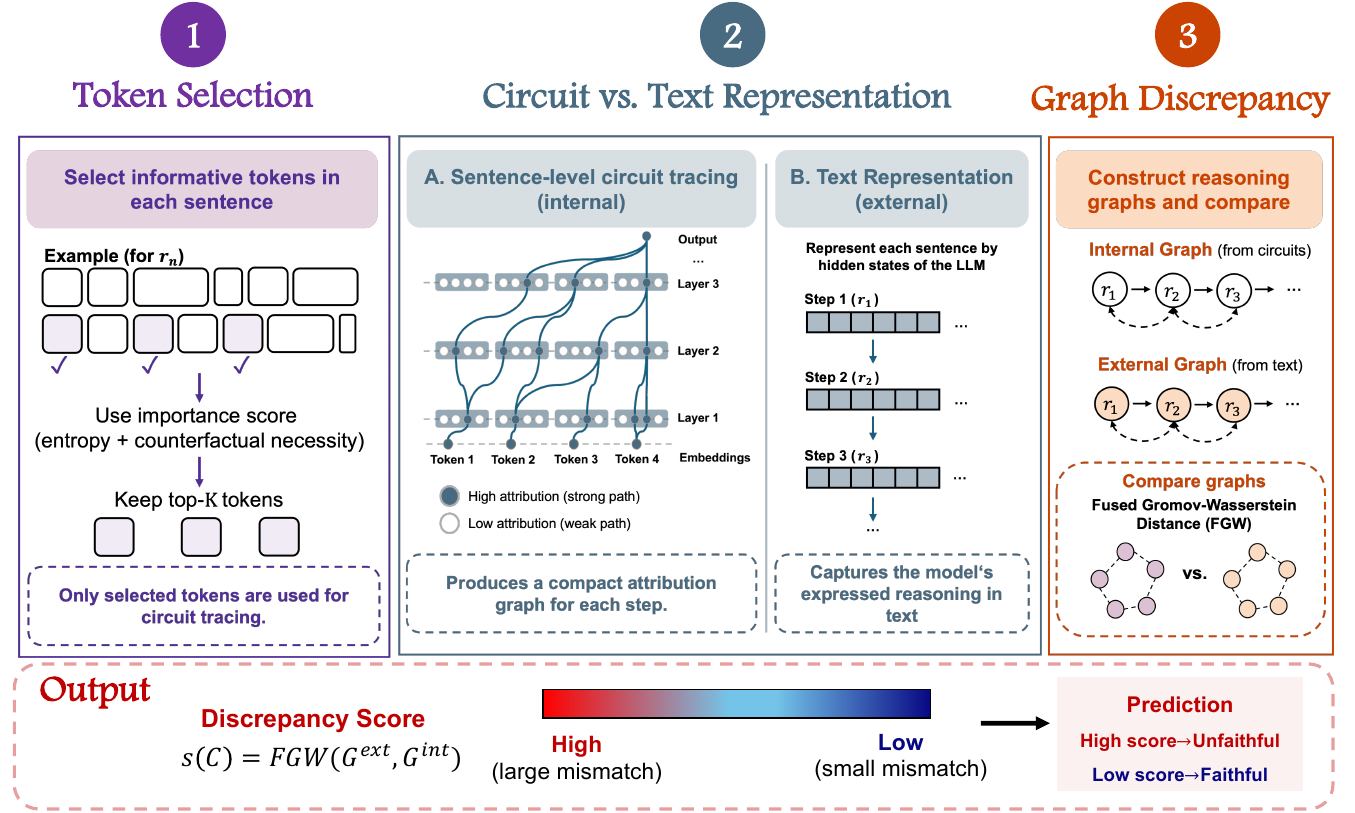}
    \caption{Overview of the \ourmethod{} for CoT unfaithfulness detection. The framework selects informative reasoning tokens and traces sentence-level internal attribution circuits from model activations. These internal circuit representations are compared against external text-based reasoning representations. The discrepancy between the two reasoning graphs is measured using FGW distance.}
    \label{fig:overview}
\end{figure}
In this section, we introduce \textbf{C}ircuit-guided \textbf{I}nternal-\textbf{E}xternal Discrepancy Scorer (\ourmethod{}) for CoT unfaithfulness detection. \ourmethod{} first constructs compact sentence-level circuit graphs by selecting informative reasoning tokens and tracing their causal interactions (Sec.~\ref{sec:step_circuit}). It then derives two representations for each reasoning trace: an \emph{internal} representation based on circuit structures and an \emph{external} representation from model hidden states. In Sec.~\ref{sec:cie_scorer}, we organize these representations into trace-level sentence graphs and measure their structural discrepancy using the Fused Gromov--Wasserstein (FGW) distance. An overview of \ourmethod{} is shown in Fig.~\ref{fig:overview}.

\subsection{Token-Selected Step Circuit Construction}
\label{sec:step_circuit}
Since CoT unfaithfulness is defined over the full reasoning trace, one could construct a single circuit for the entire CoT. However, full-trace circuit construction is computationally expensive, produces overly fine-grained token-level graphs, and is difficult to align with the displayed reasoning process, which is naturally organized as a sequence of sentences. We therefore use each reasoning sentence as the basic unit of analysis. Let the $t$-th reasoning sentence be $r_t=(o_{t,1},o_{t,2},\dots,o_{t,L_t})$, where $o_{t,\mu}$ denotes the $\mu$-th token and $L_t$ is the sentence length. Rather than tracing all tokens in $r_t$, which introduces substantial redundancy and many low-contribution nodes, we first select a compact set of informative tokens and then construct a sentence-level circuit graph around them.

\paragraph{Entropy-based candidate selection.}
We first use token-level entropy as a lightweight signal to identify positions that are more likely to participate in active reasoning. Token entropy measures the uncertainty associated with a token in a sequence and provides a simple way to characterize the model's uncertainty when producing a token. It is useful for identifying positions that may carry stronger reasoning relevance. Specifically, for token $o_{t,\mu}$, let $z_{t,\mu} \in \mathbb{R}^{V}$ denote the pre-softmax logits at position $\mu$, where $V$ is the vocabulary size and $\mu$ indexes the token position within the sentence. Let $p_{t,\mu} = \mathrm{Softmax}(z_{t,\mu} / \tau)$ denote the corresponding predictive distribution, where $\tau$ is the decoding temperature. We then compute the token-level entropy as
\begin{equation}
H_{t,\mu} = - \sum_{j=1}^{V} p_{t,\mu,j} \log p_{t,\mu,j}.
\end{equation}
High-entropy tokens mark positions where the model is less certain, and thus may carry more reasoning-relevant information. Prior work also suggests that low-entropy tokens contribute little to reasoning, while a small subset of high-entropy tokens can preserve much of the model's reasoning capability~\citep{wang2025beyond}; therefore, we first filter out extremely low-entropy tokens:
\begin{equation}
\mathcal{C}_t
=
\left\{
o_{t,\mu}
\;\middle|\;
H_{t,\mu} \ge Q_{\rho}(H_{t,1}, \dots, H_{t,L_t})
\right\},
\end{equation}
where $Q_{\rho}(\cdot)$ denotes the empirical $\rho$-quantile, which serves as the entropy threshold for filtering out low-entropy tokens.

\paragraph{Counterfactual necessity score.}
Entropy highlights positions where the model is uncertain, but uncertainty alone does not imply causal importance. Some high-entropy tokens may carry limited semantic content, such as discourse connectives, or may be weakly linked to the final reasoning outcome. We therefore refine the candidates with a counterfactual necessity score that measures how much a token contributes to the current reasoning sentence and its immediate continuation.

For each candidate token $o_{t,\mu} \in \mathcal{C}_t$, we intervene on the decoding process by replacing it with an alternative token $\tilde{o}_{t,\mu}$ and re-generating the remaining suffix of the sentence. This yields a counterfactual sentence
\begin{equation}
r_t^{(\mu \rightarrow \tilde{o})}
=
(o_{t,<\mu}, \tilde{o}_{t,\mu}, \hat{o}_{t,>\mu}),
\end{equation}
where $\hat{o}_{t,>\mu}$ denotes the regenerated suffix conditioned on the modified prefix. We then measure the effect of this intervention in two ways. The first is a sentence-level semantic change:
\begin{equation}
D^{\mathrm{sent}}_{t,\mu}
=
1 - \cos\bigl(g(r_t), g(r_t^{(\mu \rightarrow \tilde{o})})\bigr),
\end{equation}
where $g(\cdot)$ is a \textsc{BERT}-based sentence encoder. The second is a downstream predictive change:
\begin{equation}
D^{\mathrm{traj}}_{t,\mu}
=
D_{\mathrm{KL}}
\Bigl(
p^{\mathrm{down}}(\cdot \mid q, r_t)
\;\|\;
p^{\mathrm{down}}(\cdot \mid q, r_t^{(\mu \rightarrow \tilde{o})})
\Bigr),
\end{equation}
where $p^{\mathrm{down}}(\cdot)$ denotes a downstream predictive distribution conditioned on the current sentence. The two terms capture complementary aspects: the first measures how much the sentence's content changes, while the second measures how much the model's subsequent decision process changes. Combining them encourages the retained tokens to be both semantically informative and maximally influential to the current reasoning sentence.
We combine them into a counterfactual necessity score:
\begin{equation}
N_{t,\mu}
=
\lambda D^{\mathrm{sent}}_{t,\mu}
+
(1-\lambda) D^{\mathrm{traj}}_{t,\mu},
\end{equation}
where $\lambda \in [0,1]$ controls the trade-off between local semantic change and downstream predictive change. A larger $N_{t,\mu}$ indicates that the token is more necessary for the current reasoning process.
\paragraph{Selected token-based circuit construction.}
To jointly consider uncertainty and causal importance, we first rescale the entropy scores $H_{t,\mu}$ and necessity scores $N_{t,\mu}$ within the candidate set $\mathcal{C}_t$ to the range $[0,1]$, obtaining $\tilde{H}_{t,\mu}$ and $\tilde{N}_{t,\mu}$, respectively. We then define the final token importance as
\begin{equation}
I_{t,\mu}
=
\tilde{N}_{t,\mu}\bigl(1 + \beta \tilde{H}_{t,\mu}\bigr),
\qquad
o_{t,\mu} \in \mathcal{C}_t,
\end{equation}
where $\beta \ge 0$ controls how strongly entropy modulates the final importance.
We then greedily select the top-$K$ non-redundant tokens according to $I_{t,\mu}$, yielding $\mathcal{S}_t = \{o_{t,i_1}, o_{t,i_2}, \dots, o_{t,i_K}\}$. 

Instead of tracing all token positions in $r_t$, we construct a compressed sentence-level circuit conditioned on the selected tokens:
\begin{equation}
G_t^{\mathrm{circ}} = \mathrm{Trace}(q, r_t; \mathcal{S}_t).
\end{equation}
Here, $q$ and $r_t$ provide the contextual input for activation collection and attribution computation, while $\mathcal{S}_t$ specifies the selected token positions for compressed circuit construction. In practice, we use the rightmost selected token as the tracing anchor and restrict source features to $\mathcal{S}_t$, reducing graph size while preserving reasoning-relevant internal evidence.

\subsection{Circuit-Guided Internal--External Discrepancy Scorer}
\label{sec:cie_scorer}

\ourmethod detects CoT unfaithfulness by measuring the \textbf{discrepancy} between the displayed reasoning trace and a circuit-derived approximation of the model's internal computation. The displayed CoT provides the external reasoning trajectory, while sentence-level circuit graphs provide white-box evidence of the internal decision process. To make the two sources comparable, we represent both as \textit{trace-level sentence graphs}, where nodes correspond to reasoning sentences and edges encode their dependencies. Under this formulation, unfaithfulness detection becomes an \textit{internal-external graph discrepancy scoring problem}: a larger discrepancy between the external reasoning graph and the internal computation graph indicates a higher likelihood of unfaithfulness.

\paragraph{Step-level circuit sequence.}
Given the selected sentence-level circuits from Section~\ref{sec:step_circuit}, we organize them according to the CoT step order:
\begin{equation}
\mathcal{G}^{\mathrm{circ}}
=
\left(G_1^{\mathrm{circ}},G_2^{\mathrm{circ}},\dots,G_T^{\mathrm{circ}}\right),
\end{equation}
where each $G_t^{\mathrm{circ}}$ is induced by the reasoning sentence $r_t$. This ordered circuit sequence provides a step-wise white-box approximation of the model's internal computation along the full CoT trace.

Since step circuits are sparse, heterogeneous, and variable in size, direct trace-level comparison is inefficient. We use a graph neural network to encode each circuit into a compact latent representation:
\begin{equation}
x_t^{\mathrm{int}} = f_{\mathrm{GNN}}\!\left(G_t^{\mathrm{circ}}\right), 
\qquad
X^{\mathrm{int}} =
\left(
x_1^{\mathrm{int}},
x_2^{\mathrm{int}},
\dots,
x_T^{\mathrm{int}}
\right),
\end{equation}
where $x_t^{\mathrm{int}}\in\mathbb{R}^{d}$. Each embedding $x_t^{\mathrm{int}}$ serves as the internal representation of sentence $r_t$, and $X^{\mathrm{int}}$ forms the node feature matrix of the internal reasoning graph.

\paragraph{External sentence representations.}
To characterize the displayed reasoning process, we derive a sentence-level representation for each $r_t$ from the model's hidden states. We use a fixed intermediate layer $\ell$, as intermediate representations often encode task-relevant semantic and predictive information before the final output layer~\citep{li2024focus}. Specifically, we average token hidden states within each sentence:
\begin{equation}
x_t^{\mathrm{ext}}
=
\frac{1}{L_t}
\sum_{\mu=1}^{L_t}
h_{t,\mu}^{\ell},
\qquad
X^{\mathrm{ext}}
=
\left(
x_1^{\mathrm{ext}},
x_2^{\mathrm{ext}},
\dots,
x_T^{\mathrm{ext}}
\right),
\end{equation}
where $h_{t,\mu}^{\ell}$ denotes the hidden state of the $\mu$-th token in sentence $r_t$ at layer $\ell$, and $L_t$ is the token length of $r_t$. To make external representations comparable with internal circuit embeddings, we use a lightweight MLP adaptor $P$ to project $X^{\mathrm{ext}}$ into the same $d$-dimensional space, denoted as $\tilde{X}^{\mathrm{ext}}=P(X^{\mathrm{ext}})$. The adapted sentence embeddings $\tilde{X}^{\mathrm{ext}}$ then serve as the node feature matrix of the external reasoning graph.

\paragraph{Shared sentence-graph construction.}
Given sentence representations $X=(x_1,x_2,\dots,x_T)$, we construct a trace-level graph by treating $X$ as the node feature matrix. The same construction rule is applied to both adapted external representations $\tilde{X}^{\mathrm{ext}}$ and internal representations $X^{\mathrm{int}}$, yielding $G^{\mathrm{ext}}=(A^{\mathrm{ext}},\tilde{X}^{\mathrm{ext}})$ and $G^{\mathrm{int}}=(A^{\mathrm{int}},X^{\mathrm{int}})$. This shared rule ensures that the final discrepancy score reflects differences between displayed and internal reasoning structures rather than artifacts of graph construction. The adjacency matrix captures two complementary aspects of a reasoning trace: \textbf{local temporal progression} and \textbf{long-range semantic dependency}. Specifically, each sentence is connected to its immediate successor to preserve the sequential flow of the CoT, while additional forward edges are weighted by the positive semantic similarity between sentence representations:
\begin{equation}
A_{ij}
=
\begin{cases}
\mathbb{I}[j=i+1]+\lambda\max\!\left(0,\cos(x_i,x_j)\right), & i<j,\\
0, & j\le i,
\end{cases}
\end{equation}
where $\lambda$ controls the strength of semantic dependency edges. We row-normalize $A$ to stabilize the scale of outgoing dependencies, and set the node measure $\mu$ to be uniform over reasoning sentences. This construction provides a unified graph abstraction for both displayed and internal reasoning: node features encode sentence-level reasoning states, while edges encode temporal and semantic dependencies across the trace.

\paragraph{FGW-based unfaithfulness score.}
After the shared graph construction, we obtain an external reasoning graph 
$G^{\mathrm{ext}}=(A^{\mathrm{ext}},\tilde{X}^{\mathrm{ext}})$ and an internal reasoning graph 
$G^{\mathrm{int}}=(A^{\mathrm{int}},X^{\mathrm{int}})$. Since CoT unfaithfulness reflects a mismatch between the displayed reasoning trace and the internal computation that supports the answer, the discrepancy between these two graphs provides direct evidence for unfaithfulness. We compare the two graphs using the \textbf{Fused Gromov--Wasserstein (FGW) distance}, which measures graph matching by jointly considering \textit{node-feature discrepancy} and \textit{structural discrepancy}: the former captures whether each displayed reasoning step is consistent with its internal computational evidence, while the latter captures whether dependencies among displayed steps match the internal information-flow structure. For two graphs $G_1=(A_1,X_1,\mu_1)$ and $G_2=(A_2,X_2,\mu_2)$, the FGW distance is defined as
\begin{equation}
\begin{aligned}
\mathrm{FGW}_{\alpha}(G_1,G_2)
=
\min_{\pi\in\Pi(\mu_1,\mu_2)}
\sum_{i,j,k,l}
\Big[
\alpha \big(A_1(i,j)-A_2(k,l)\big)^2
+
(1-\alpha)\|X_1(i)-X_2(k)\|_2^2
\Big]\pi_{ik}\pi_{jl},
\end{aligned}
\end{equation}
where $\pi$ denotes a soft node correspondence between the two graphs, $\Pi(\mu_1,\mu_2)$ is the set of admissible couplings between node measures, and $\alpha\in[0,1]$ controls the trade-off between structural and feature-level discrepancies. In practice, $\mu^{\mathrm{ext}}$ and $\mu^{\mathrm{int}}$ are set as uniform measures over reasoning sentences. We define the instance-level unfaithfulness score as $s(\mathcal{C})=\mathrm{FGW}_{\alpha}(G^{\mathrm{ext}},G^{\mathrm{int}})$, and instantiate the binary detector $f(q,\mathcal{C})$ by thresholding this score. A larger $s(\mathcal{C})$ indicates a stronger internal--external graph discrepancy, suggesting a higher likelihood of CoT unfaithfulness.

\paragraph{Training objective.}
Let $y_n \in \{0,1\}$ denote the trace-level label of the $n$-th sample, where $y_n=1$ indicates an unfaithful trace and $y_n=0$ indicates a faithful trace. Since faithful traces should exhibit smaller internal--external graph discrepancy, while unfaithful traces should exhibit larger discrepancy, we train the model with a margin-based distance objective:
\begin{equation}
\mathcal{L}
=
\frac{1}{N}
\sum_{n=1}^{N}
\left[
(1-y_n) \, s(\mathcal{C}_n)
+
y_n \, \max\!\bigl(0, m - s(\mathcal{C}_n)\bigr)
\right],
\end{equation}
where $m>0$ is a margin specifying the minimum desired discrepancy for unfaithful traces. The first term minimizes FGW distance for faithful samples, encouraging consistency between external and internal graphs, while the second term pushes unfaithful samples to have a discrepancy larger than $m$. During inference, a trace is predicted as unfaithful when its score exceeds a validation-set threshold. The pseudo-code algorithm of
\ourmethod is provided in Appendix~\ref{sec:pseudocode}
\section{Experiments}
\label{sec:exp}
\subsection{Experimental Setup}
\paragraph{Dataset.} We evaluate our method on \texttt{FaithCoT-Bench} proposed by \citet{shen2025faithcot}, a benchmark designed for instance-level CoT unfaithfulness detection. The benchmark consists of CoT trajectories collected from four domain-specific datasets, with human annotations assigned according to predefined unfaithfulness criteria, mainly including \emph{post-hoc reasoning} and \emph{spurious reasoning chains}. Specifically, it includes \texttt{LogicQA}~\citep{liu2020logiqa}, \texttt{TruthfulQA}~\citep{truthfulqa}, \texttt{AuQA}~\citep{auqa}, and the biomedical subset of \texttt{HLE}~\citep{HLE}. These datasets cover four representative domains: logical, factual, mathematical, and biological reasoning. Detailed statistics are provided in Appendix~\ref{app:dataset}.
\begin{table*}[t]
\centering
\caption{Comparison of CoT faithfulness detection methods on four datasets. Results are reported in Acc and F1 (\%). The best result for each metric under each dataset is highlighted in \textbf{bold}. ``OOM'' indicates that the method runs out of memory.}
\label{tab:llama31_main}
\setlength{\tabcolsep}{3pt}
\renewcommand{\arraystretch}{0.9}
\small
\resizebox{0.9\textwidth}{!}{%
\begin{tabular}{llcccccccc}
\toprule
\multirow{2}{*}{\textbf{Paradigm}} & \multirow{2}{*}{\textbf{Method}} 
& \multicolumn{2}{c}{\textbf{Logic-QA}} 
& \multicolumn{2}{c}{\textbf{Truthful-QA}} 
& \multicolumn{2}{c}{\textbf{AQuA}} 
& \multicolumn{2}{c}{\textbf{HLE-Bio}} \\
\cmidrule(lr){3-4} \cmidrule(lr){5-6} \cmidrule(lr){7-8} \cmidrule(lr){9-10}
& 
& Acc$\uparrow$ & F1$\uparrow$
& Acc$\uparrow$ & F1$\uparrow$
& Acc$\uparrow$ & F1$\uparrow$
& Acc$\uparrow$ & F1$\uparrow$ \\
\midrule

\multirow{2}{*}{\textit{Baselines}}
& Random            & 42.0 & 35.4 & 43.3 & 42.7 & 43.2 & 37.4 & 35.7 & 43.8 \\
& Perplexity        & 52.3 & 19.2 & 44.4 & 40.5 & 49.4 & 36.1 & 47.4 & 52.4 \\
\midrule

\multirow{5}{*}{\shortstack[l]{\textit{Counterfactual}\textit{-based}}}
& Adding Mistakes   & 57.5 & 47.9 & 51.1 & 60.7 & 71.6 & 66.7 & 46.4 & 51.6 \\
& Option Shuffling  & 48.3 & 52.6 & 51.1 & 59.3 & 61.0 & 16.7 & 40.0 & 14.3 \\
& Removing Steps    & 51.7 & 27.6 & 36.7 & 50.4 & 72.4 & 46.2 & 39.3 & 37.0 \\
& Early Answering   & 35.2 & 48.6 & 40.0 & 52.6 & 72.4 & 53.3 & 46.4 & 48.3 \\
& Paraphrasing      & 42.5 & 47.9 & 67.8 & 49.1 & 68.4 & 42.9 & 45.7 & 40.0 \\
\midrule

\multirow{2}{*}{\shortstack[l]{\textit{Logits}\textit{-based}}}
& Answer Tracing    & 33.0 & 45.9 & 38.9 & 50.5 & 53.2 & 30.8 & 64.3 & 76.2 \\
& Information Gain  & 51.7 & 51.2 & 47.8 & 40.5 & 41.6 & 20.2 & 52.5 & 9.5 \\
\midrule

\multirow{1}{*}{\shortstack[l]{\textit{LLM-as-}\textit{judge}}}
& BiGGen     & 52.9 & 59.4 & 61.1 & 67.3 & 73.7 & {70.3} & 70.2 & 69.2 \\
\midrule
\multirow{2}{*}{\shortstack[l]{\textit{Circuit}\textit{-based}}}
& CRV    & - & -& 68.0 & 66.7 & 63.0 & 62.8 & - & - \\
& \ourmethod & \textbf{69.0} & \textbf{60.8} & \textbf{78.0} & \textbf{71.5} & \textbf{77.0} & \textbf{72.8} & \textbf{78.0} & \textbf{79.7} \\
\bottomrule
\end{tabular}%
}
\vspace{-0.1in}
\end{table*}
\paragraph{Baselines.}
We compare our method with 11 representative CoT faithfulness detection baselines, grouped into 5 categories. First, \emph{simple baselines} include a random classifier as a lower-bound reference and a perplexity-based method~\citep{perplexity} that treats sentence fluency as an indicator of faithfulness. Second, \emph{counterfactual-based methods} evaluate whether modifying the CoT changes the final answer, including Adding Mistakes, Option Shuffling, Removing Steps, Early Answering, and Paraphrasing~\citep{lanham2023measuring}. Third, \emph{logit-based methods}, such as Answer Tracing~\citep{llm_trustworthiness} and Information Gain~\citep{information_gain}, detect unfaithfulness by analyzing token-level probability dynamics. Fourth, \emph{LLM-as-Judge methods} use BiGGen~\citep{biggen}, a rubric-based evaluation framework, to assess reasoning faithfulness with fine-grained criteria. 
Finally, \emph{circuit-based methods}, represented by CRV~\citep{crv}, construct attribution circuits from model computations and extract interpretable circuit-level features for downstream faithfulness classification. Detailed introduction provided in Appendix~\ref{app:baseline}.
\paragraph{Implementation}
For baselines, we follow the official implementation and evaluation protocol of FaithCoT-Bench to ensure fair comparison. For the CRV-based baseline, we use the Gradient Boosting Classifier. We implement circuit tracing using the \texttt{crv-8b-instruct-transcoders} released by CRV, which is matched to the backbone model used for CoT generation, i.e., \texttt{Llama-3.1-8B-Instruct}. For external sentence representations, we extract hidden states from the 15-th layer. For the graph encoder, we use a 2-layer GIN with hidden and output dimension 256. The external adaptor is a two-layer MLP with same dimension. All experiments are conducted on a single NVIDIA A100 GPU with 80GB memory. Detailed hyperparameters provided in Appendix~\ref{app:hyperparameters}.

\subsection{Performance comparison }
\textbf{\ourmethod{} achieves state-of-the-art performance across all datasets and metrics.}
As shown in Table~\ref{tab:llama31_main}, \ourmethod{} obtains the best Acc and F1 on all four datasets using the LLaMA3.1 backbone. Compared with the strongest previous result for each metric, \ourmethod{} improves Acc by 11.5, 10.0, 5.4, and 7.8 points on Logic-QA, Truthful-QA, AQuA, and HLE-Bio, respectively, and improves F1 by 1.4, 4.2, 2.5, and 3.5 points. These consistent gains demonstrate the effectiveness of our trace-level graph alignment framework for CoT faithfulness detection. 

\textbf{Existing baselines provide useful but incomplete faithfulness signals.}
Simple baselines such as Random and Perplexity perform poorly, suggesting that surface-level fluency is insufficient for identifying unfaithful reasoning. Counterfactual-based methods are competitive on some datasets, but their performance varies substantially across tasks, since external perturbations may not directly reflect the internal causes of unfaithfulness. Logits-based methods capture token-level probability changes, yet they are also unstable across datasets. The rubric-based LLM-as-judge method BiGGen is a strong baseline, but it relies mainly on the generated reasoning text and does not explicitly access internal computation. Compared with CRV, \ourmethod{} is more scalable: CRV fails on Logic-QA and HLE-Bio due to dense full-circuit tracing over all token positions, whereas our compressed circuit construction successfully handles all datasets while preserving key internal reasoning signals.

\subsection{Ablation study}
\begin{wraptable}{r}{0.62\textwidth}
\vspace{-1em}
\centering
\caption{Ablation study on four datasets. Results are reported in Acc and F1 (\%). The best result for each metric under each dataset is highlighted in \textbf{bold}.}
\label{tab:ablation_study}
\setlength{\tabcolsep}{2.5pt}
\renewcommand{\arraystretch}{1.0}
\scriptsize
\resizebox{0.62\textwidth}{!}{%
\begin{tabular}{lcccccccc}
\toprule
\multirow{2}{*}{\textbf{Variant}}
& \multicolumn{2}{c}{\textbf{Logic-QA}}
& \multicolumn{2}{c}{\textbf{Truthful-QA}}
& \multicolumn{2}{c}{\textbf{HLE-Bio}}
& \multicolumn{2}{c}{\textbf{AQuA}} \\
\cmidrule(lr){2-3}
\cmidrule(lr){4-5}
\cmidrule(lr){6-7}
\cmidrule(lr){8-9}
& Acc & F1 & Acc & F1 & Acc & F1 & Acc & F1 \\
\midrule

\ourmethod
& \textbf{69.0} & \textbf{60.8}
& \textbf{78.0} & \textbf{71.5}
& \textbf{77.0} & \textbf{72.2}
& \textbf{78.0} & \textbf{79.7} \\

Entropy only
& 58.0 & 59.8
& 70.0 & 57.3
& 75.0 & 62.2
& 59.0 & 52.9 \\

Counterfactual only
& 61.0 & 58.7
& 68.0 & 62.5
& 74.0 & 70.7
& 73.0 & 57.1 \\

w/o sequential edges
& 63.0 & 51.8
& 75.0 & 66.7
& 69.0 & 65.4
& 75.0 & 64.2 \\

w/o similarity edges
& 65.0 & 53.7
& 74.0 & 68.1
& 73.0 & 65.0
& 69.0 & 70.3 \\

w/o internal GNN
& 60.0 & 58.3
& 67.0 & 63.1
& 64.0 & 65.7
& 71.0 & 58.6 \\

\bottomrule
\end{tabular}%
}
\vspace{-5pt}
\end{wraptable}
\textbf{Token selection is critical for identifying internal reasoning signals.}
When replacing our token selection strategy with single variants, the performance consistently drops across all datasets as shown in Table~\ref{tab:ablation_study}. Compared with the full model, the entropy-only variant leads to an average drop of about 15.3\% in Acc and 15.5\% in F1, with particularly clear degradation on AQuA. The counterfactual-only variant yields an average decrease of about 10.3\% in Acc and 10.7\% in F1. These results indicate that single-signal selection cannot reliably preserve the key internal evidence for faithfulness detection. \textbf{By integrating complementary token-level signals, our token selection reduces circuit-tracing cost while producing more informative internal reasoning graphs.}

\textbf{Graph structure and internal graph encoding are both necessary for detection.}
As shown in Table~\ref{tab:ablation_study}, removing sequential or similarity edges consistently degrades performance. Specifically, sequential-edge ablation causing an average drop of about 9.6\% in Acc and 12.7\% in F1, and the similarity-edge ablation causing an average drop of about 8.7\% in Acc and 9.0\% in F1. This shows that reasoning order and semantic proximity provide complementary information for modeling circuit. Removing the internal GNN further leads to larger degradation, with an average drop of about 14.7\% in Acc and 14.3\% in F1. \textbf{These results suggest that our method benefits not only from selected circuit features, but also from explicitly modeling their graph structure.}
\begin{figure*}[t]
\centering
\begin{subfigure}[t]{0.47\textwidth}
    \centering
    \includegraphics[width=\linewidth,height=0.45\linewidth]{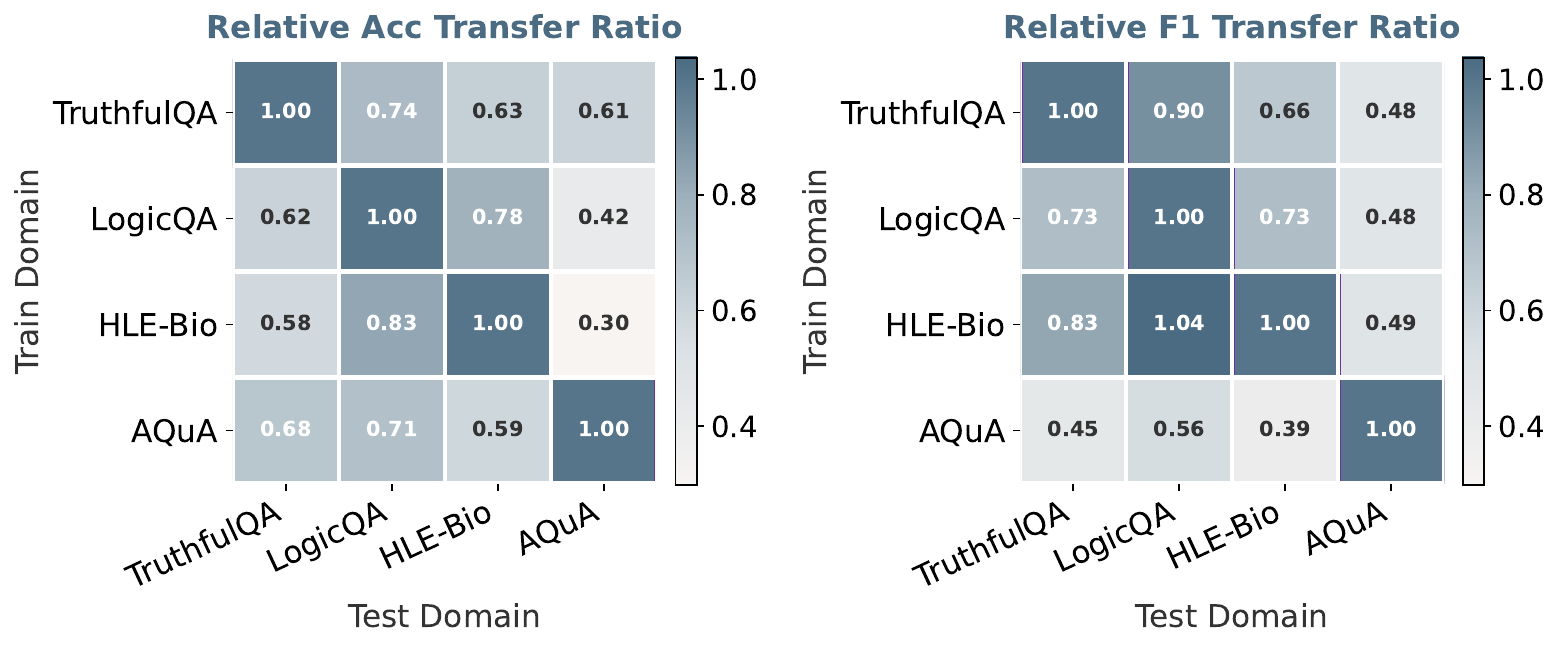}
    \caption{Cross-domain transfer.}
    \label{fig:cross_domain_rtr}
\end{subfigure}
\hfill
\begin{subfigure}[t]{0.45\textwidth}
    \centering
    \includegraphics[width=\linewidth,height=0.5\linewidth]{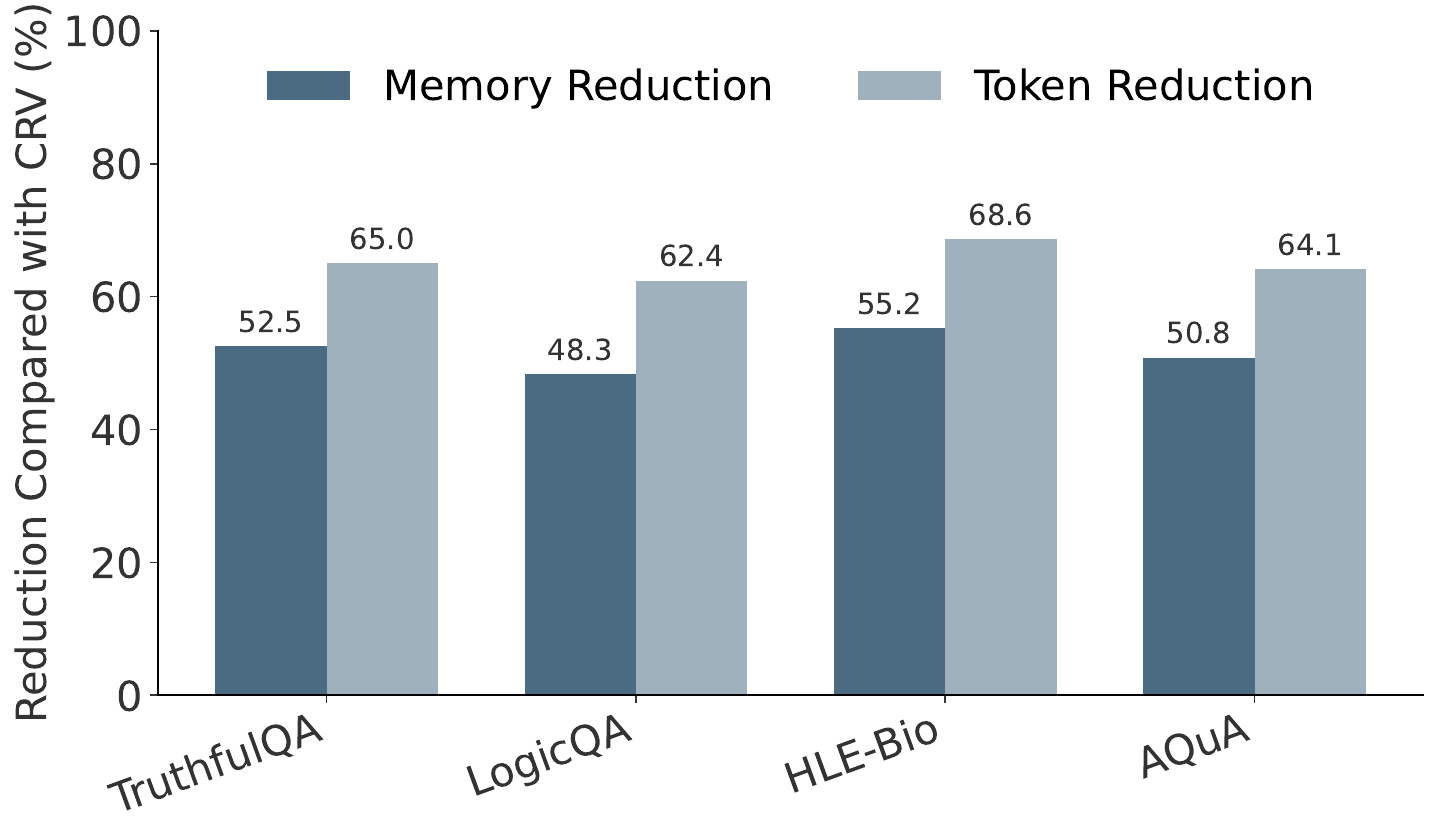}
    \caption{Efficiency analysis.}
    \label{fig:efficiency_reduction}
\end{subfigure}
\caption{Additional analysis of \ourmethod. 
(a) Cross-domain generalization measured by relative transfer ratio.
(b) Memory and token reduction compared with CRV.
}
\vspace{-15pt}
\label{fig:additional_analysis}
\end{figure*}
\vspace{-5pt}
\subsection{In-depth analysis}
\begin{wrapfigure}{r}{0.45\textwidth}
\vspace{-25pt}
 % 上移整个浮动体
\centering
\includegraphics[width=\linewidth]{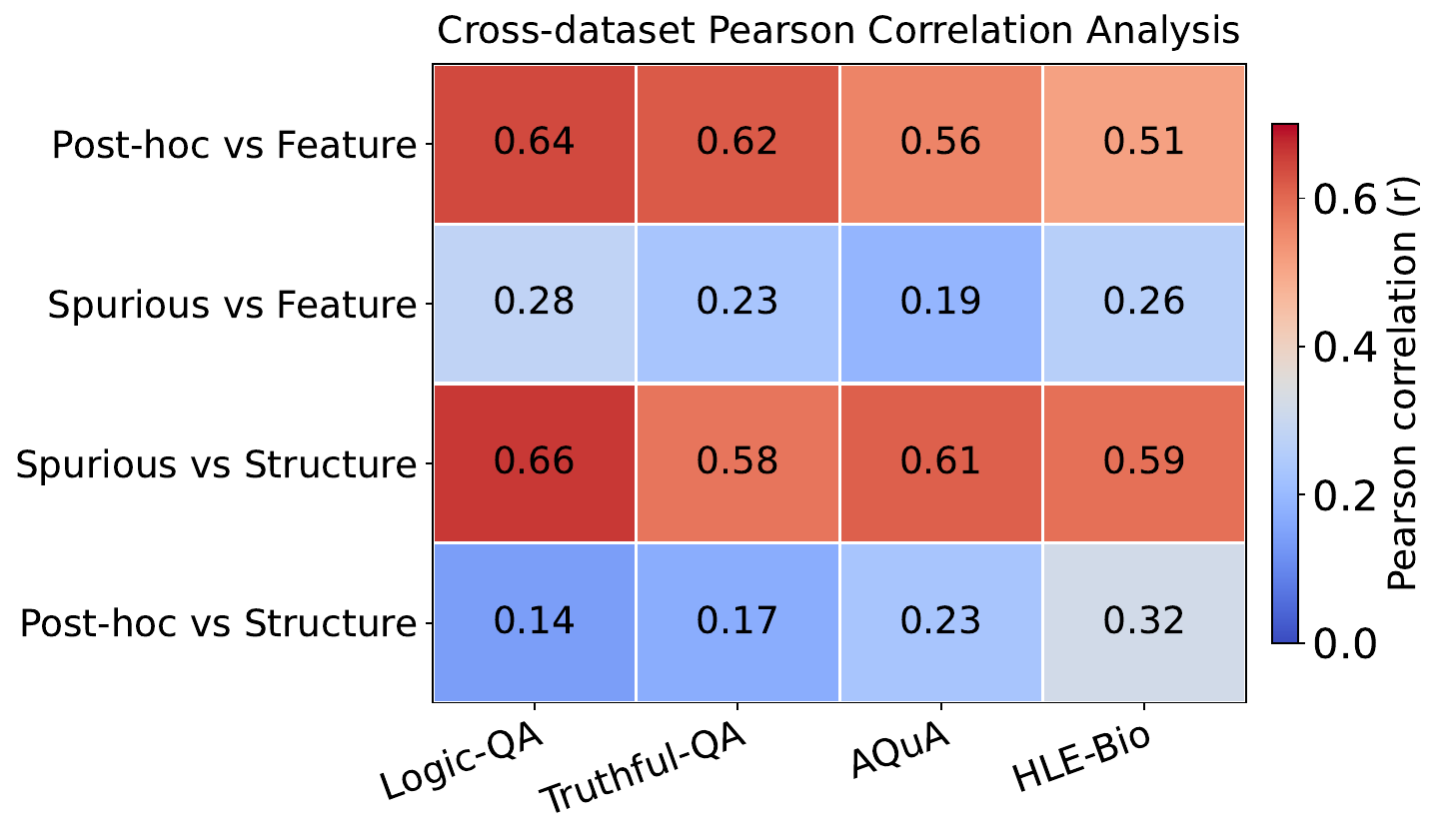}
 % 缩小图和 caption 之间距离
 \vspace{-10pt}
\caption{
Cross-dataset Pearson correlation 
}
\vspace{-30pt}
\label{fig:type_corr}
\end{wrapfigure}
\textbf{Different unfaithfulness types exhibit distinct internal--external mismatch patterns.}
Following the~\citep{shen2025faithcot} taxonomy, we divide unfaithful CoTs into post-hoc reasoning and spurious reasoning chains.
Based on the FGW objective, we decompose the discrepancy score into its feature-level term 
$s_{\mathrm{feat}}=\sum_{i,j,k,l}\|\tilde{X}^{\mathrm{ext}}(i)-X^{\mathrm{int}}(k)\|_2^2$ 
and structure-level term 
$s_{\mathrm{struct}}=\sum_{i,j,k,l}(A^{\mathrm{ext}}(i,j)-A^{\mathrm{int}}(k,l))^2$, 
and compute Pearson correlations between each type indicator and the two components.
Post-hoc reasoning tends to induce stronger feature-level discrepancy, as its displayed steps are constructed to rationalize a predetermined answer and may lack support from internal circuit evidence. By contrast, spurious reasoning chains mainly induce structure-level discrepancy, since their locally coherent steps may break causal dependencies with the question or final answer. As shown in Figure~\ref{fig:type_corr}, post-hoc reasoning correlates more with feature-level than structure-level discrepancy ($0.58$ vs. $0.22$), while spurious reasoning chains show the opposite trend ($0.61$ vs. $0.24$ for structure vs. feature). We provide more detailed analysis in Appendix~\ref{app:type_analysis} and case study in Appendix~\ref{app:case}.

\subsection{Cross-domain Performance}
\textbf{\ourmethod demonstrates meaningful cross-domain transfer across several reasoning domains.}
To evaluate cross-domain generalization, we train the detector on one dataset and directly test it on the remaining datasets as shown in Figure~\ref{fig:cross_domain_rtr}. We report the relative transfer ratio (RTR), defined as $\mathrm{RTR}_{i \rightarrow j} = \mathrm{Perf}(\mathrm{train}=i,\mathrm{test}=j) / \mathrm{Perf}(\mathrm{train}=j,\mathrm{test}=j)$, where $\mathrm{Perf}$ denotes either Acc or F1. Thus, diagonal entries are normalized to $1.0$, and a higher off-diagonal value indicates stronger transfer to the target domain. The results show that \ourmethod transfers well on several domain pairs: TruthfulQA achieves an F1 RTR of $0.905$ on Logic-QA, while HLE-Bio obtains $1.036$ on Logic-QA and $0.825$ on TruthfulQA. This suggests that the learned internal graph representation captures faithfulness-related signals that can generalize across domains. Meanwhile, transfer involving AQuA is weaker, with most off-diagonal F1 RTR values below $0.5$, indicating a larger domain gap for mathematical reasoning traces.
\subsection{Efficiency Analysis}
\textbf{\ourmethod achieves better scalability than CRV by reducing both tracing cost and computational resources.}
As shown in Figure~\ref{fig:efficiency_reduction}, \ourmethod consistently reduces peak memory usage by 48.3\%--55.2\% and traced tokens by 62.4\%--68.6\% across the four datasets. This indicates that the proposed token selection strategy can effectively filter out redundant tracing targets before circuit construction, thereby reducing both the size of extracted circuits and the memory required for attribution computation. Moreover, on the datasets where CRV can successfully run, \ourmethod achieves clear runtime reductions of 67.6\% on TruthfulQA and 46.7\% on AQuA, while also improving F1 by 7.2\% and 15.9\%, respectively, as shown in Figure~\ref{fig:efficiency_runtime_f1}. These results suggest that the efficiency gain does not come from simply discarding useful evidence; instead, by focusing circuit tracing on more informative tokens, \ourmethod obtains a better efficiency--effectiveness trade-off than CRV.
% \textbf{\ourmethod improves scalability over CRV by reducing tracing cost and computational overhead.} As shown in Figure~\ref{fig:efficiency_reduction}, \ourmethod reduces peak memory usage by 48.3\%--55.2\% and traced tokens by 62.4\%--68.6\% across the four datasets. This confirms that our token selection strategy effectively removes redundant tracing targets and makes circuit construction more lightweight. Moreover, on the datasets where CRV can successfully run, \ourmethod achieves clear runtime reductions of 67.6\% on TruthfulQA and 46.7\% on AQuA, while also improving F1 by 7.2\% and 15.9\%, respectively, as shown in Figure~\ref{fig:efficiency_runtime_f1}. These results demonstrate that \ourmethod improves computational efficiency without sacrificing detection performance.
% \midrule
% \multirow{2}{*}{Feature/structure source}
% & Feature only 
% & 65.0 & 59.8 
% & 77.0 & 69.3 
% & 75.6 & 72.2 
% & 88.0 & 70.0 \\

% & Structure only 
% & 60.0 & 48.7 
% & 78.0 & 72.5 
% & \textbf{78.0} & 72.7 
% & 73.0 & 57.1 \\
\vspace{-5pt}
\section{Conclusion}
\vspace{-5pt}
In this paper, we propose \ourmethod, a circuit-guided framework for instance-level CoT unfaithfulness detection that measures the discrepancy between internal computational evidence and displayed reasoning trajectories. \ourmethod selects informative tokens to construct compact sentence-level circuit graphs, encodes them with a GNN into internal reasoning embeddings, and aligns them with external reasoning graphs via FGW distance to produce an unfaithfulness score. Experiments on four FaithCoT-Bench datasets show that \ourmethod achieves state-of-the-art performance, demonstrating the effectiveness and efficiency of the proposed framework.

\bibliographystyle{plainnat}
\bibliography{reference}

%%%%%%%%%%%%%%%%%%%%%%%%%%%%%%%%%%%%%%%%%%%%%%%%%%%%%%%%%%%%
\clearpage
\appendix
\section{Related Works}
\label{app:rw}
\subsection{Understanding Chain-of-Thought Reasoning}
Chain-of-thought (CoT) reasoning has become a widely adopted mechanism for enhancing the reasoning ability of large language models (LLMs)~\citep{cot,zero_cot,tot,got,hong1,hong2,hong3,hong4}. 
By eliciting intermediate steps before the final answer, CoT improves performance on complex tasks such as mathematics, programming, logic, commonsense, and scientific reasoning~\citep{math_intro,code_intro,logic_intro,common_intro}. 
Beyond its original use as a prompting strategy, CoT traces have increasingly been used as supervision signals for reasoning-oriented training, reinforcement learning, distillation, and compression, enabling models to acquire or transfer multi-step reasoning abilities more effectively~\citep{long_cot,yang2024survey,chen2025towards,shen2025understanding,hou2025thinkprune}. 
Thus, CoT has evolved from a prompting technique into a central paradigm for both improving reasoning performance and exposing intermediate model behavior. 
However, generated reasoning traces may not faithfully reflect the model's actual decision process, motivating the need to examine CoT as an object of faithfulness analysis.

\subsection{Faithfulness Evaluation for Chain-of-Thought}

CoT traces are often treated as natural-language explanations of LLM decisions, but recent studies show that such apparent transparency can be misleading~\citep{feng2023towards,zhao2025chain,not_explainability,matton2025walk}. 
Existing work mainly evaluates CoT faithfulness from two perspectives. 
The first studies the causal role of CoT by intervening on reasoning traces, such as adding mistakes, removing steps, paraphrasing rationales, or shuffling options, and observing whether the final prediction changes~\citep{lanham2023measuring,turpin2023language,yang2025well,xiong2025measuring}. 
The second uses LLM-as-Judge methods or process reward models to assess step quality, factual correctness, or answer consistency~\citep{wild_faithful,step_judge}. 
However, these approaches either measure whether CoT affects the answer at a general level or rely mainly on external textual rationales, making it difficult to determine whether a specific generated trace faithfully reflects the model's internal decision process. 
In contrast, our work targets instance-level CoT unfaithfulness detection by explicitly comparing displayed reasoning with evidence from the model's internal computation.

\subsection{Mechanistic Interpretability}
Mechanistic interpretability aims to explain model behavior by analyzing the internal computations that produce predictions~\citep{zhao2024explainability,du2019techniques}. 
Recent sparse autoencoder (SAE) studies decompose dense activations into sparse, human-interpretable features, while circuit analysis examines how these features interact and causally contribute to model outputs~\citep{gaoscaling,hannahave,conmy2023towards,cunningham2023sparse}. 
These techniques move beyond black-box input-output behavior and provide a more direct view of the computational evidence used by the model. 
Circuit tracing further constructs attribution graphs over internal features, offering white-box evidence for whether a displayed CoT aligns with the model's computational trajectory~\citep{ameisen2025circuit,elhage2021mathematical}. 
This makes circuit-level evidence particularly relevant for CoT faithfulness evaluation, where the key question is whether the written rationale reflects the computation behind the answer. 
However, applying full circuit tracing to long multi-step CoTs is computationally expensive, and recent circuit-based reasoning analysis still faces this scalability issue~\citep{crv}. 
\section{Dataset}
\label{app:dataset}
We use FaithCoT-Bench~\citep{shen2025faithcot}, a benchmark for instance-level CoT faithfulness evaluation. The benchmark is constructed from four source datasets spanning different reasoning domains:

\begin{itemize}
    \item \texttt{LogicQA}~\citep{liu2020logiqa} originates from the Chinese National Civil Servants Examination, later translated into English, and is specifically designed to test \textit{logical deductive reasoning}. Each question is paired with a short passage and multiple candidate answers, requiring the model to select the option that logically follows. Unlike general reading comprehension tasks, \texttt{LogicQA} emphasizes multi-step deduction and resistance to spurious correlations, making it a strong testbed for evaluating the causal structure of reasoning traces.  
    \item \texttt{TruthfulQA}~\citep{truthfulqa} consists of factual questions drawn from diverse domains, deliberately crafted to elicit common misconceptions or human-like falsehoods. Presented in multiple-choice format, the benchmark requires models to reject plausible-sounding but false answers and instead provide factually correct responses. This adversarial setup directly tests whether reasoning faithfully distinguishes truth from widely held but misleading beliefs.
    \item \texttt{AQuA}~\citep{auqa} evaluates \textit{step-by-step numerical reasoning} through algebraic word problems. Each instance requires decomposing the problem into intermediate reasoning steps that lead to a numerical solution. Because arithmetic reasoning depends on tightly coupled causal chains, this dataset provides a natural setting for examining whether CoTs maintain faithful intermediate justifications rather than skipping or fabricating steps. 
    \item The biomedical portion of \texttt{HLE}~\citep{HLE}, referred to as \texttt{HLE-Bio}, represents \textit{knowledge-intensive biomedical reasoning}. Questions are drawn from technical biomedical texts and require the integration of domain-specific knowledge with logical inference. Compared to the other benchmarks, \texttt{HLE-Bio} poses heightened challenges due to its specialized terminology and dense factual grounding, offering a critical stress test for the faithfulness of reasoning traces.
\end{itemize}
Together, these source datasets allow FaithCoT-Bench to cover logical, factual, mathematical, and biomedical reasoning, making it suitable for evaluating CoT faithfulness across diverse task settings.

\section{Baselines}
\label{app:baseline}

To comprehensively evaluate CoT unfaithfulness detection, we compare \ourmethod with representative baselines from four categories: counterfactual-based methods, logit-based methods, LLM-as-Judge methods, and circuit-based methods.

\ding{192} \textbf{Counterfactual-based methods.}
Counterfactual-based methods assess CoT faithfulness by perturbing the generated reasoning trajectory and measuring whether the final answer changes. Given an original CoT $\mathcal{C}$ and its answer $a$, a counterfactual operator $\Delta$ produces a perturbed trajectory $\mathcal{C}'=\Delta(\mathcal{C})$, from which the model generates a new answer $a'=\mathrm{M}_{\text{Ans}}(q,\mathcal{C}')$. If $a'\ne a$, the perturbed reasoning content is considered causally influential to the prediction. Conversely, if the answer remains unchanged after substantial perturbation, the modified steps may not be part of the causal reasoning path, suggesting potential unfaithfulness.

We evaluate the following representative perturbation strategies:
\begin{itemize}
    \item \textit{Adding Mistakes}: Inserts misleading or incorrect intermediate reasoning steps and tests whether the final answer is affected.
    \item \textit{Option Shuffling}: Randomizes the order of multiple-choice options to examine whether the reasoning and prediction remain stable under changed option contexts.
    \item \textit{Removing Steps}: Deletes selected reasoning steps to evaluate whether the removed content is necessary for deriving the answer.
    \item \textit{Early Answering}: Forces the model to answer before completing the full reasoning trajectory, testing whether later steps are causally needed.
    \item \textit{Paraphrasing}: Rewrites reasoning steps while preserving or slightly altering their semantics, measuring whether the model's prediction is robust to surface-level or semantic variations.
\end{itemize}
These methods are widely used to probe whether a CoT causally contributes to the final prediction~\citep{lanham2023measuring,yang2025well,xiong2025measuring,paul2024making,yee2024dissociation,yee2024dissociation}.

\ding{193} \textbf{Logit-based methods.}
Logit-based methods analyze the model's probability dynamics during reasoning. The key assumption is that a faithful reasoning trace should provide evidence that gradually supports the final answer. At each generation step, the model outputs a logit vector $\mathbf{z}_i$ over the vocabulary, which is converted into a token probability by
\begin{align}
p(c_i \mid \cdot)
= \mathrm{softmax}(\mathbf{z}_i)[c_i]
= \frac{\exp(z_{i,c_i})}{\sum\nolimits_k \exp(z_{i,k})}.
\end{align}

\noindent\textbf{Answer Tracing.}
Answer Tracing measures how the probability of the final answer changes as more reasoning tokens are generated. Let $a$ denote the final answer token and $C_{1:i}=(c_1,\dots,c_i)$ denote the prefix of the reasoning trace. The step-wise confidence change is computed as
\begin{align}
\Delta_i = p(a \mid C_{1:i}) - p(a \mid C_{1:i-1}), \quad i=1,\dots,T.
\end{align}
A faithful CoT is expected to provide increasing support for the final answer as reasoning unfolds, whereas inconsistent or weakly supportive probability trends may indicate unfaithfulness.

\noindent\textbf{Information Gain.}
Information Gain measures how much the question reduces uncertainty over the generated reasoning trace. Given a question $Q$ and a CoT $C=(c_1,\dots,c_n)$, it is defined as
\begin{align}
IG(C, Q) &= H(C) - H(C \mid Q) \\ \nonumber
&= - \sum_{i=1}^{n} p(c_i \mid C_{1:i-1}) \log p(c_i \mid C_{1:i-1}) 
+ \sum_{i=1}^{n} p(c_i \mid C_{1:i-1}, Q) \log p(c_i \mid C_{1:i-1}, Q).
\end{align}
A higher information gain suggests that the question provides stronger guidance for generating the reasoning trace, which is treated as a proxy signal for faithfulness. Overall, logit-based methods use internal probability signals, but they do not explicitly model the circuit-level computational structure behind the prediction.

\ding{194} \textbf{LLM-as-Judge methods.}
LLM-as-Judge methods use a stronger language model to evaluate whether a generated CoT is faithful. Following the fine-grained rubric-based evaluation protocol of BiGGen~\citep{biggen}, we provide the judge model with the question, generated CoT, final answer, and detailed scoring criteria, and ask it to produce a faithfulness judgment. Compared with direct prompting, this rubric-guided setting encourages the judge to assess the reasoning trace in a more structured manner, considering aspects such as logical consistency, support for the final answer, and potential post-hoc rationalization. Although such methods can capture fine-grained textual patterns, they still mainly rely on the external content of generated rationales and may conflate reasoning correctness, plausibility, and faithfulness.

\ding{195} \textbf{Circuit-based method.}
We also compare with CRV~\citep{crv}, a circuit-based method that uses mechanistic interpretability signals for reasoning analysis. For each CoT instance or reasoning step, CRV constructs a circuit graph to represent the model's internal computation. The graph is then converted into a fixed-dimensional feature vector using a set of handcrafted graph statistics, which is finally fed into a classifier $F$ for prediction:
\begin{align}
    \mathbf{x}_{\mathrm{CRV}} = \phi(\mathcal{G}_{\mathrm{circuit}}), 
    \quad
    \hat{y} = F(\mathbf{x}_{\mathrm{CRV}}),
\end{align}
where $\mathcal{G}_{\mathrm{circuit}}$ denotes the constructed circuit graph, $\phi(\cdot)$ denotes the graph feature extraction function, and $F$ is the downstream classifier.

CRV extracts handcrafted features from each circuit graph and feeds them into a classifier. These features summarize the graph from three aspects: global statistics such as active node count and output uncertainty, node-level activation and influence statistics, and structural properties such as density, connectivity, and centrality. Although these features provide useful circuit-level signals, CRV mainly treats each circuit as an individual graph with summary statistics. In contrast, \ourmethod models the discrepancy between internal computational trajectories and external reasoning trajectories, while using token selection to reduce circuit-tracing cost.
\section{Pseudo-code for \ourmethod}
\label{sec:pseudocode}
Algorithm~\ref{alg:training} summarizes the training procedure of \ourmethod. 
Given a CoT trace, CIE-SCORER first constructs compact sentence-level circuits by selecting informative tokens and tracing their internal attribution structures. 
It then encodes these circuits into internal sentence representations, extracts external sentence representations from hidden states, and builds two trace-level sentence graphs under the same graph construction rule. 
Finally, the model is trained by minimizing a margin-based objective over the FGW distance between the internal and external graphs, encouraging faithful traces to have small discrepancies and unfaithful traces to have large discrepancies.
\begin{algorithm}[t]
\caption{Training Procedure of \ourmethod}
\label{alg:training}
\KwIn{
Training set $\mathcal{D}=\{(q_n,\mathcal{C}_n,y_n)\}_{n=1}^{N}$;
target LLM $\pi_\theta$; tracing budget $K$;
hyperparameters $\rho,\lambda,\beta,\alpha,m$.
}
\KwOut{Trained graph encoder $f_{\mathrm{GNN}}$ and adaptor $P$.}

Initialize $f_{\mathrm{GNN}}$ and adaptor $P$\;

\ForEach{epoch}{
\ForEach{mini-batch $\mathcal{B}\subset\mathcal{D}$}{
    $\mathcal{L}_{\mathcal{B}}\leftarrow 0$\;
    
    \ForEach{$(q,\mathcal{C},y)\in\mathcal{B}$}{
        \ForEach{reasoning sentence $r_t\in\mathcal{C}$}{
            Select informative tokens 
            $
            \mathcal{S}_t
            =
            \mathrm{TokenSelect}(q,r_t;\rho,\lambda,\beta,K)
            $\;
            
            Construct compressed circuit graph 
            $
            G_t^{\mathrm{circ}}
            =
            \mathrm{Trace}(q,r_t;\mathcal{S}_t)
            $\;
            
            Encode internal representation
            $
            x_t^{\mathrm{int}}
            =
            f_{\mathrm{GNN}}(G_t^{\mathrm{circ}})
            $\;
            
            Extract external representation
            $
            x_t^{\mathrm{ext}}
            =
            \mathrm{MeanPool}
            (\{h_{t,\mu}^{\ell}\}_{\mu=1}^{L_t})
            $\;
        }
        
        Construct internal sentence graph 
        $
        G^{\mathrm{int}}
        =
        \mathrm{BuildGraph}(X^{\mathrm{int}})
        $\;
        
        Construct external sentence graph
        $
        G^{\mathrm{ext}}
        =
        \mathrm{BuildGraph}(P(X^{\mathrm{ext}}))
        $\;
        
        Compute discrepancy score
        $
        s(\mathcal{C})
        =
        \mathrm{FGW}_{\alpha}
        (G^{\mathrm{ext}},G^{\mathrm{int}})
        $\;
        
        Compute training loss
        $
        \mathcal{L}
        =
        (1-y)s(\mathcal{C})
        +
        y\max(0,m-s(\mathcal{C}))
        $\;
        
        $\mathcal{L}_{\mathcal{B}}
        \leftarrow
        \mathcal{L}_{\mathcal{B}}+\mathcal{L}$\;
    }
    
    Update $f_{\mathrm{GNN}}$ and $P$ by minimizing 
    $
    \mathcal{L}_{\mathcal{B}}/|\mathcal{B}|
    $\;
}
}
\Return{$f_{\mathrm{GNN}}$, $P$}\;
\end{algorithm}
\begin{table}[t]
\centering
\caption{Hyper-parameter configuration on four datasets.}
\label{tab:hyperparams}
\setlength{\tabcolsep}{8pt}
\renewcommand{\arraystretch}{1.15}
\small
\begin{tabular}{lccccccc}
\toprule
\textbf{Dataset} 
& $\lambda$ 
& $\beta$ 
& $\alpha$ 
& \textbf{GNN layers} 
& \textbf{GNN dim} 
& \textbf{Batch size} 
& \textbf{Learning rate} \\
\midrule
Logic-QA    & 0.6 & 0.5 & 0.4 & 2 & 256 & 8 & 1e-4 \\
Truthful-QA & 0.5 & 0.5 & 0.5 & 2 & 256 & 10 & 1e-4\\
AQuA        & 0.5 & 0.5 & 0.5 & 2 & 256 & 10 & 1e-4 \\
HLE-Bio     & 0.6 & 0.5 & 0.5 & 1 & 128 & 10 & 5e-5 \\
\bottomrule
\end{tabular}
\end{table}
\section{Hyperparameter}
\label{app:hyperparameters}
We report the main hyperparameter settings used in our experiments in Table~\ref{tab:hyperparams}. These hyperparameters are selected on the validation set and kept fixed during testing for each dataset.
\section{Additional experimental results}
\subsection{Hyperparameter Analysis}
\label{app:hyperparameter}
\begin{figure}[t]
  \centering
  % 第一行：两张并排
  \subfloat[Efficiency-F1]{%
    \label{fig:efficiency_runtime_f1}%
    \includegraphics[
      width=0.34\columnwidth,height=2.9cm
    ]{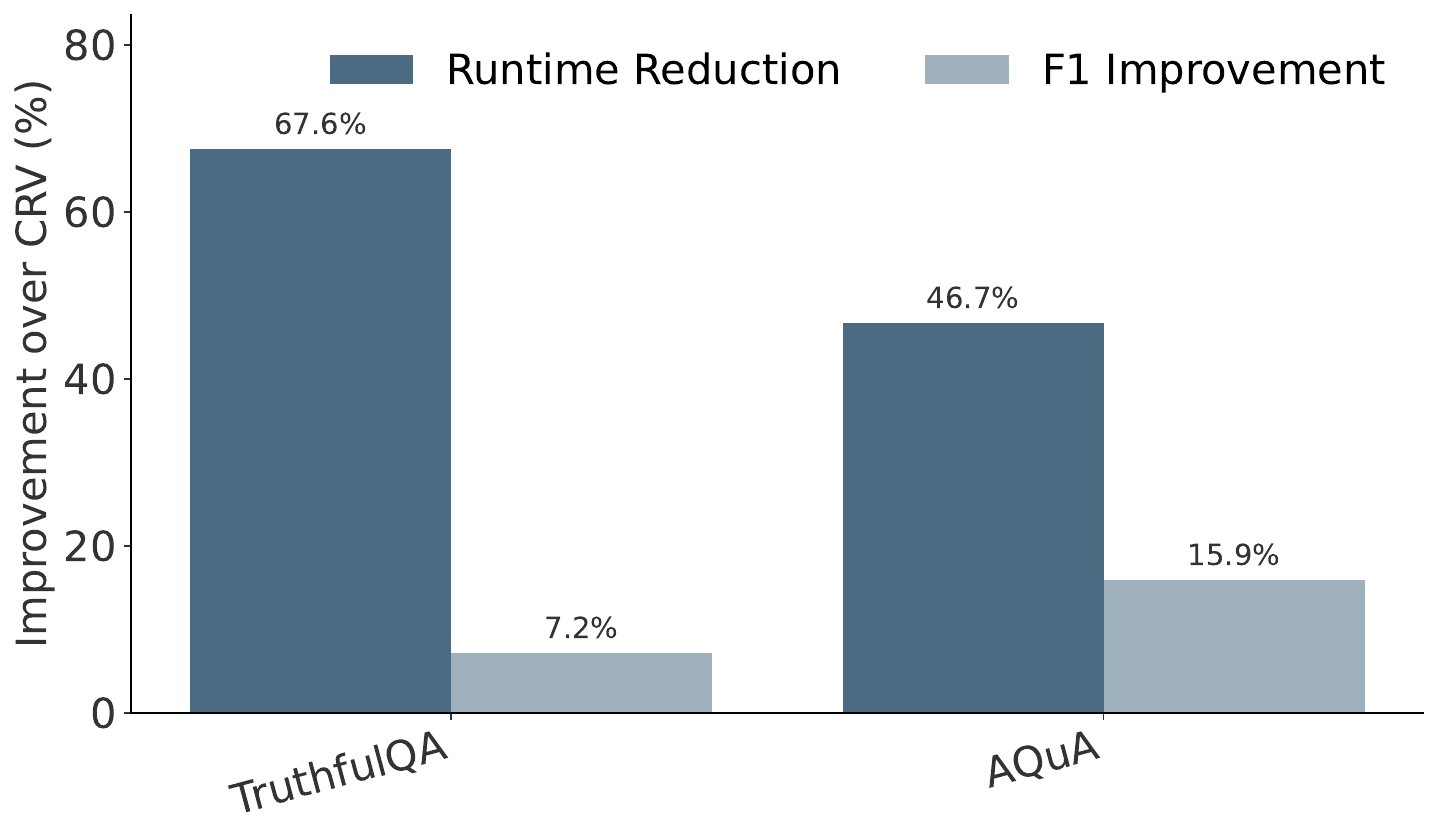}%
  }\hfill
  \subfloat[$\lambda$ Sensitivity]{%
    \label{fig:lambda_sensitivity}%
    \includegraphics[
      width=0.32\columnwidth,height=2.9cm
    ]{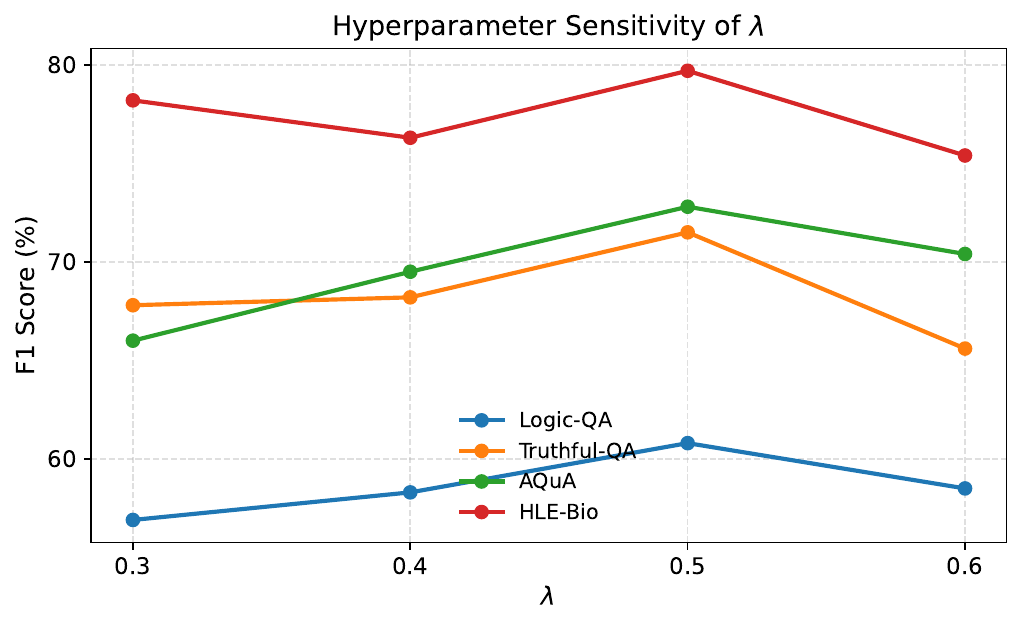}%
  }\hfill
  \subfloat[$\beta$ Sensitivity]{%
    \label{fig:beta_sensitivity}%
    \includegraphics[
      width=0.32\columnwidth,height=2.9cm
    ]{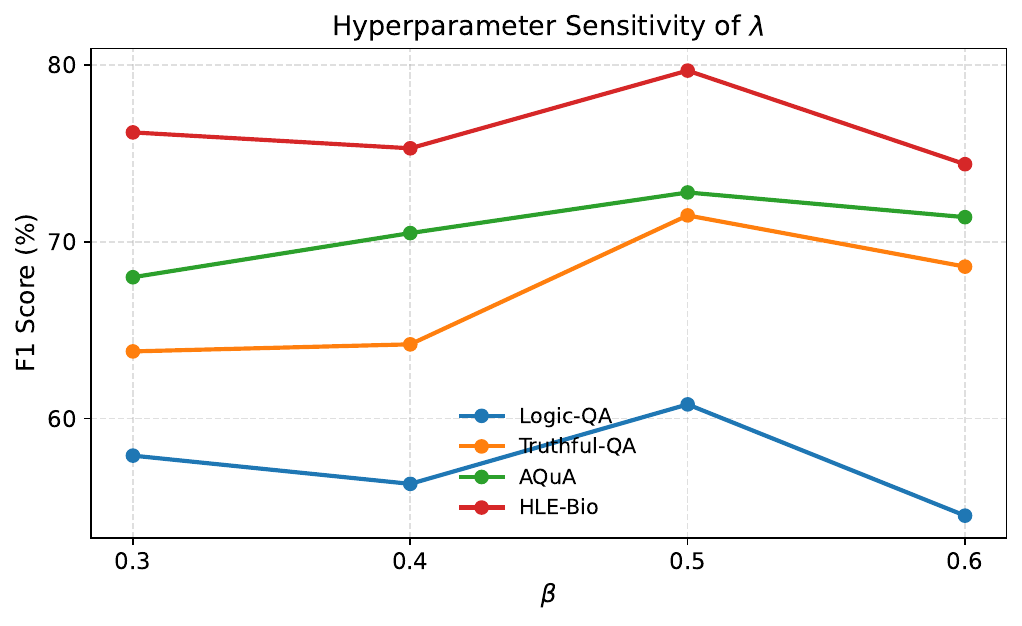}%
  }
  \caption{(a) \ourmethod reduces runtime while improving F1 on the datasets where CRV can be successfully executed. 
(b)--(c) Sensitivity analysis of the token-selection hyperparameters $\lambda$ and $\beta$. 
}
  \vspace{-6mm}
\end{figure}

\textbf{\ourmethod benefits from balanced token-selection signals.}
We analyze two token-selection hyperparameters: $\lambda$, which balances local semantic change and downstream predictive change in the counterfactual necessity score, and $\beta$, which controls the entropy modulation strength in token importance.
As shown in Figure~\ref{fig:lambda_sensitivity} and Figure~\ref{fig:beta_sensitivity}, the default setting $0.5$ achieves the best F1 score across all four datasets for both hyperparameters.
Performance decreases when either parameter deviates from $0.5$, suggesting that relying too heavily on a single signal leads to less effective token selection.
Overall, these results show that a balanced combination of semantic change, downstream influence, and entropy uncertainty is important for constructing informative sentence-level circuits.
\label{app:exp}
\subsection{In-depth analysis}
\label{app:type_analysis}

\textbf{Different unfaithfulness types exhibit distinct discrepancy patterns across datasets.}
To further validate the motivation of \ourmethod, we provide an in-depth type-wise analysis of the internal--external discrepancy. 
Following the benchmark taxonomy, we divide CoTs into three categories: faithful CoTs, post-hoc reasoning, and spurious reasoning chains. 
For each category, we report the overall FGW discrepancy together with its two decomposed components: feature-level discrepancy and structure-level discrepancy.

\begin{figure*}[t]
\centering

\begin{subfigure}{0.48\textwidth}
    \centering
    \includegraphics[width=\linewidth]{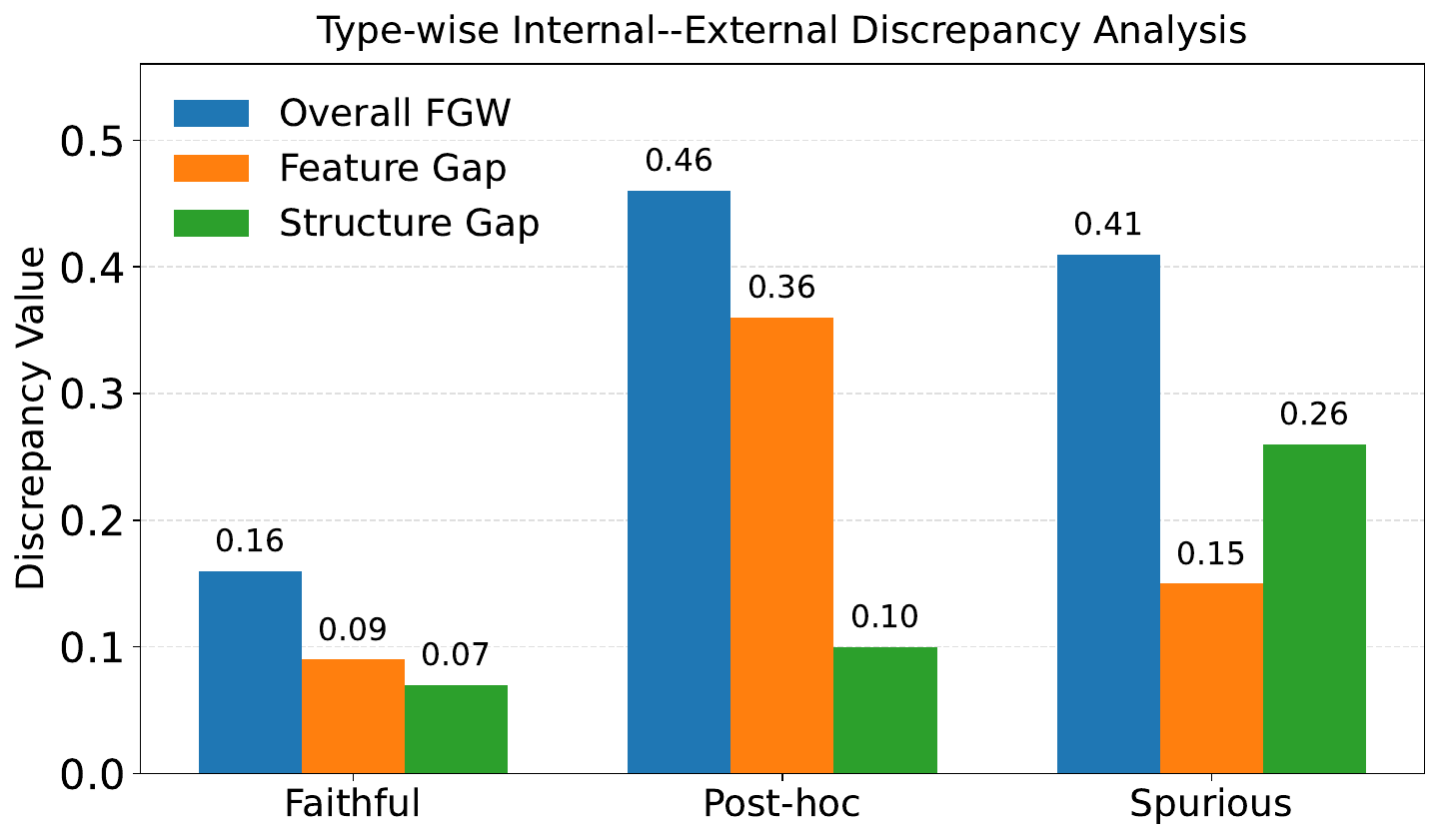}
    \caption{Logic-QA}
    \label{fig:app_type_logicqa}
\end{subfigure}
\hfill
\begin{subfigure}{0.48\textwidth}
    \centering
    \includegraphics[width=\linewidth]{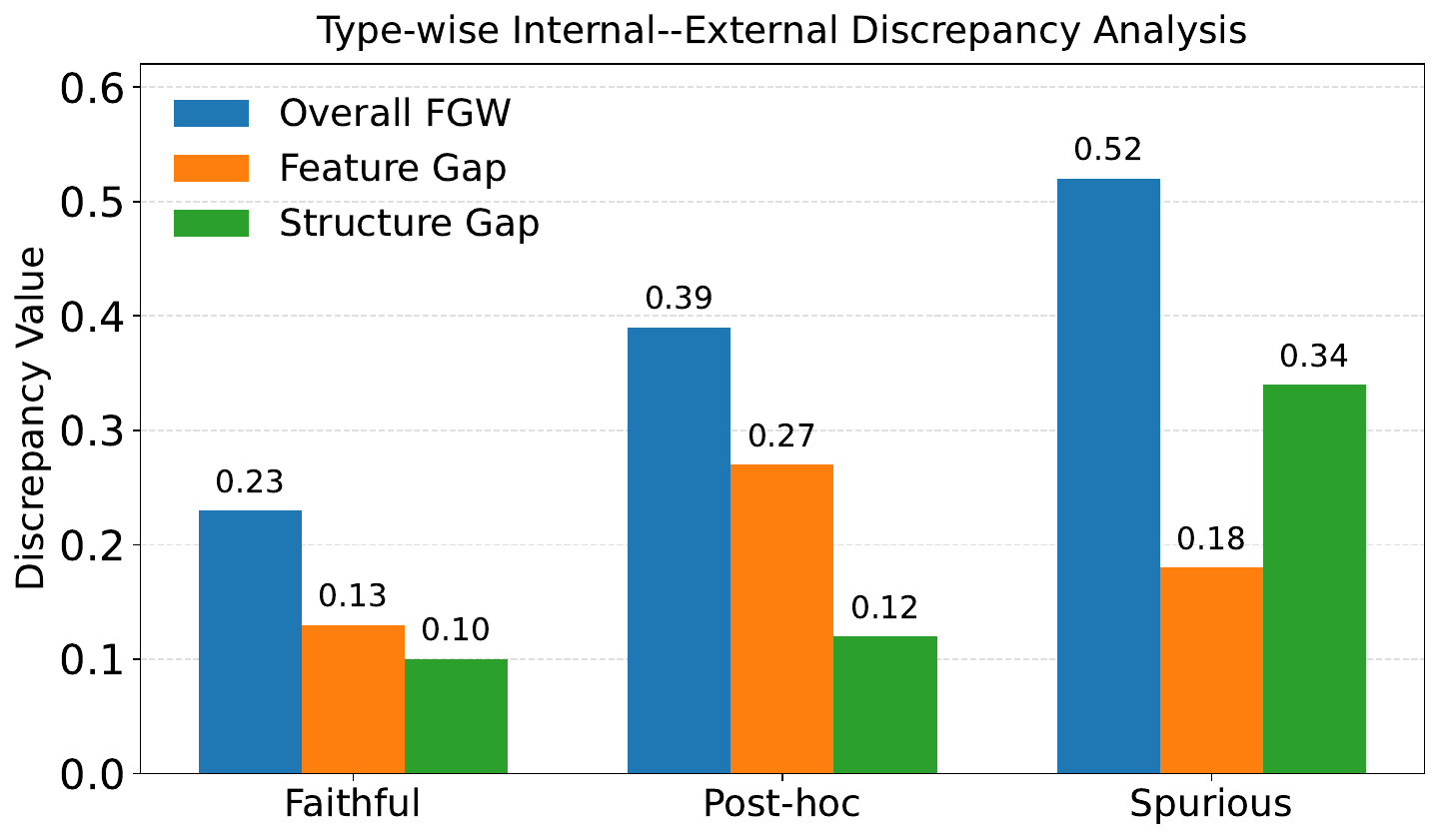}
    \caption{Truthful-QA}
    \label{fig:app_type_truthfulqa}
\end{subfigure}

\vspace{0.6em}

\begin{subfigure}{0.48\textwidth}
    \centering
    \includegraphics[width=\linewidth]{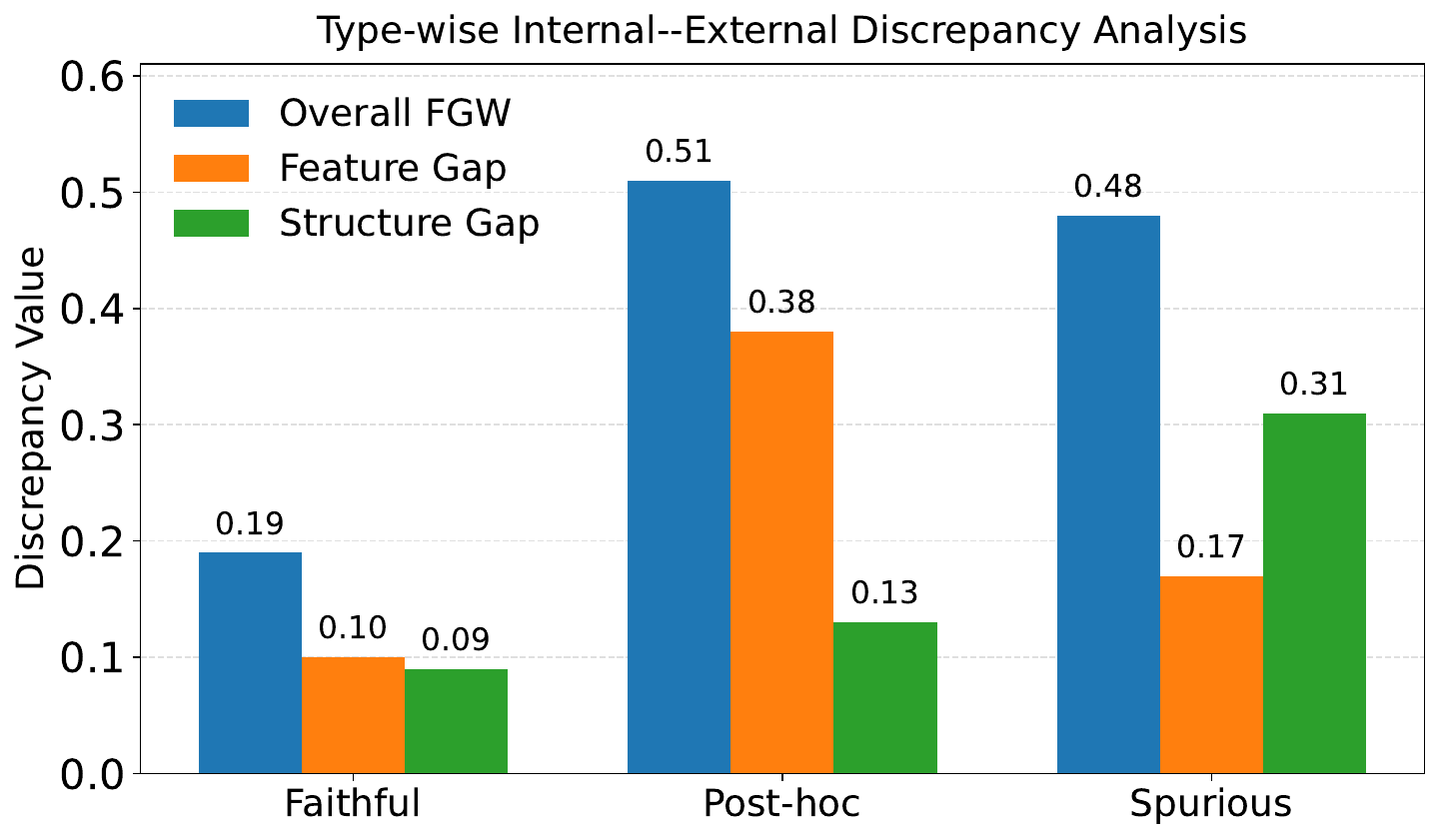}
    \caption{AQuA}
    \label{fig:app_type_aqua}
\end{subfigure}
\hfill
\begin{subfigure}{0.48\textwidth}
    \centering
    \includegraphics[width=\linewidth]{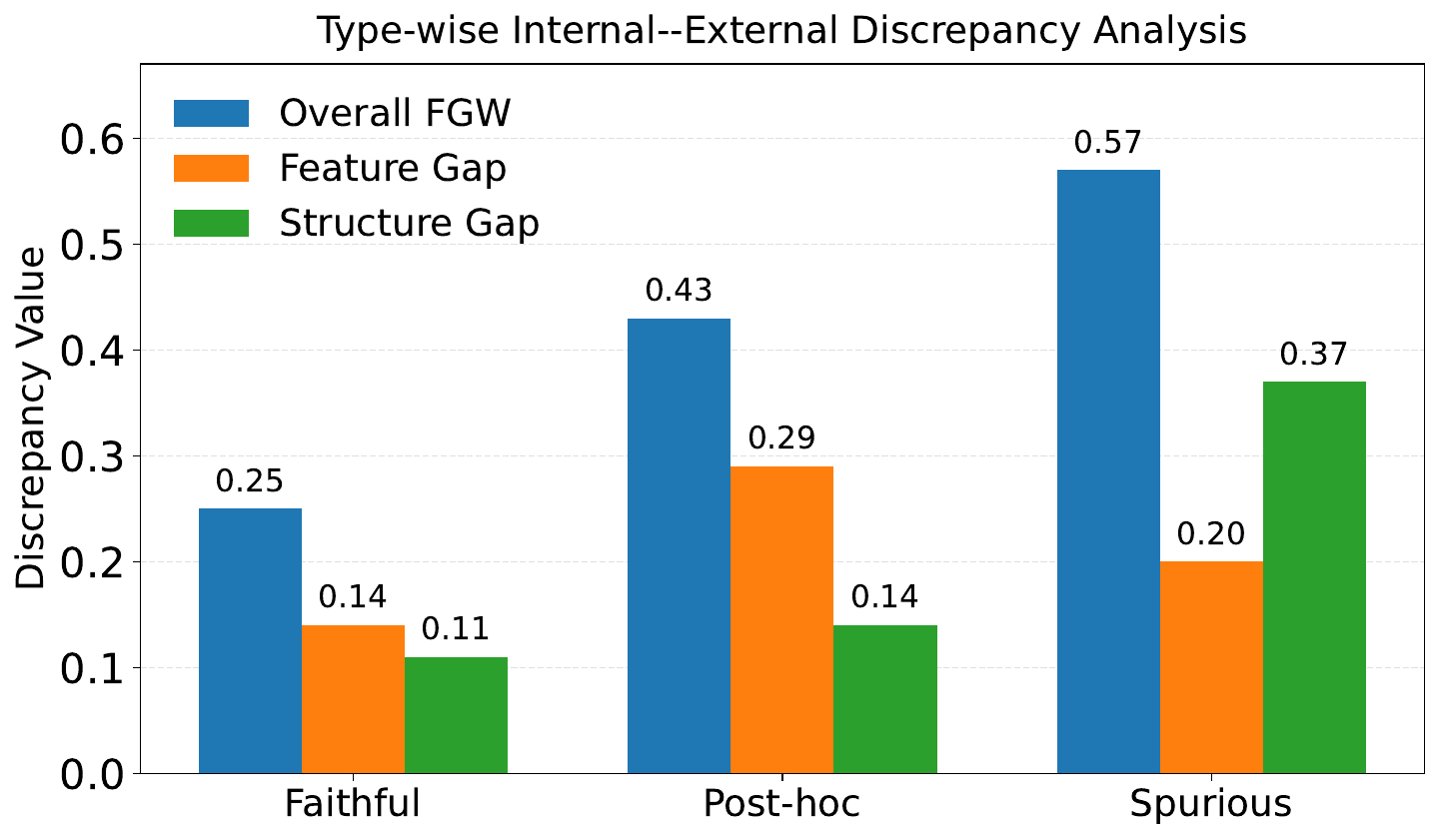}
    \caption{HLE-Bio}
    \label{fig:app_type_hlebio}
\end{subfigure}

\caption{
Type-wise internal--external discrepancy analysis across four datasets. 
Each subfigure reports the overall FGW discrepancy and its feature-level and structure-level components for faithful CoTs, post-hoc reasoning, and spurious reasoning chains. 
}
\label{fig:app_type_discrepancy}
\end{figure*}

As shown in Figure~\ref{fig:app_type_discrepancy}, faithful CoTs consistently exhibit the lowest overall discrepancy across all datasets, indicating better alignment between the displayed reasoning trace and the internal circuit-derived computation. 
For post-hoc reasoning, the feature-level discrepancy is substantially larger than the structure-level discrepancy, e.g., $0.32$ vs. $0.10$ on Logic-QA and $0.33$ vs. $0.11$ on Truthful-QA. 
This supports the intuition that post-hoc CoTs are mainly problematic because their displayed reasoning content is not well supported by the corresponding internal circuit evidence. 
In contrast, spurious reasoning chains show the opposite pattern: their structure-level discrepancy dominates the feature-level discrepancy, e.g., $0.35$ vs. $0.14$ on AQuA and $0.30$ vs. $0.16$ on HLE-Bio. 
This suggests that spurious chains mainly fail by breaking the dependency structure between reasoning steps, even when individual steps appear locally coherent. 
Overall, these results provide additional evidence that different forms of CoT unfaithfulness correspond to different internal--external mismatch patterns.
\begin{figure}[t]
    \centering
    \includegraphics[width=0.95\linewidth]{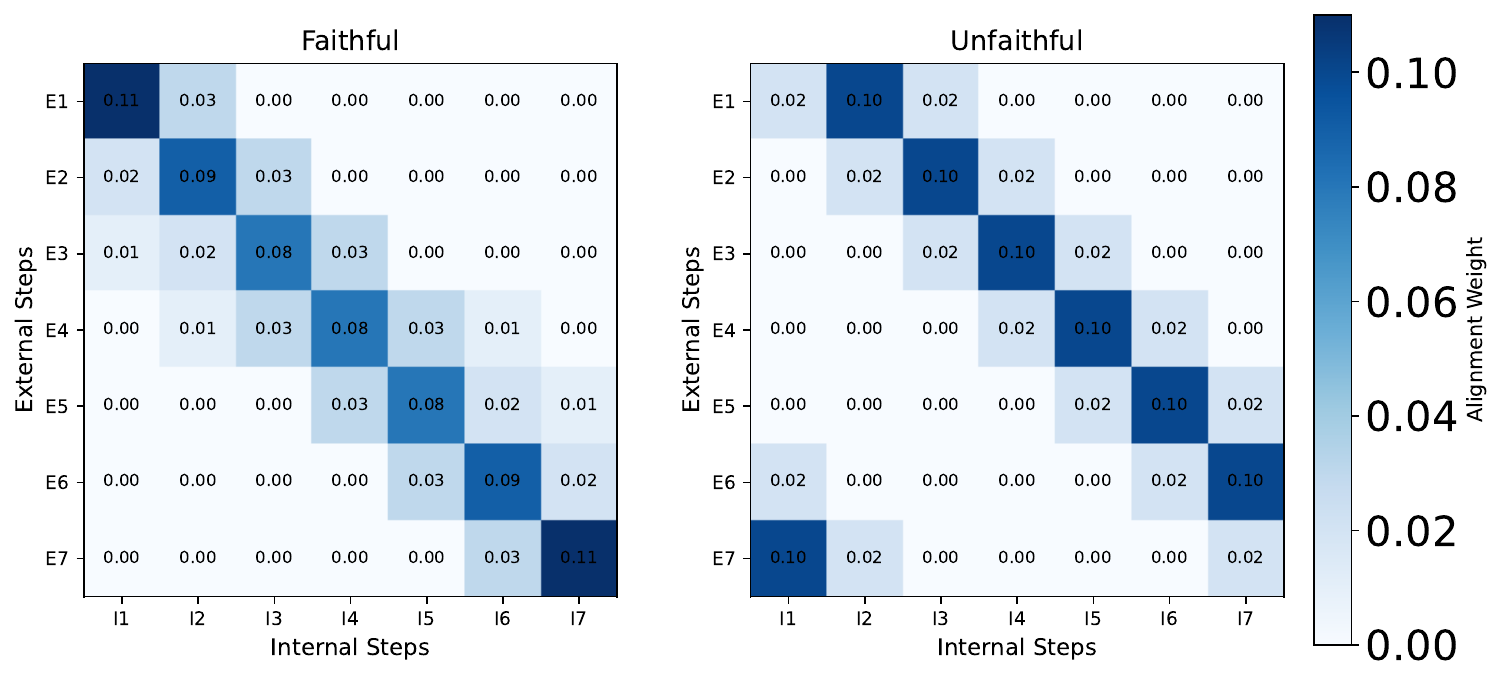}
    \caption{
    Case study of FGW coupling matrices.
    The faithful example shows near-diagonal alignment between external and internal steps, while the unfaithful example exhibits more off-diagonal alignment.
    }
    \label{fig:case_alignment}
\end{figure}
\subsection{Case study of alignment patterns.}
\label{app:case}
We visualize the FGW coupling matrix $\pi$ for representative faithful and unfaithful CoTs.
This visualization helps illustrate how the external reasoning steps are matched to the internal circuit-derived steps.
As shown in Figure~\ref{fig:case_alignment}, the faithful example exhibits a near-diagonal alignment pattern, indicating that each displayed reasoning step is mainly supported by its corresponding internal computation step.
In contrast, the unfaithful example shows a more shifted and diffuse alignment pattern, suggesting weaker and less consistent correspondence between the displayed reasoning trajectory and the internal computation process.
These examples provide an intuitive view of why internal--external discrepancy is informative for unfaithfulness detection.
\section{Limitation}
\label{app:limitation}
While \ourmethod achieves promising results, it requires access to model internals and is therefore mainly applicable to white-box or open-source LLMs. n this work, we focus on textual CoT traces from FaithCoT-Bench, leaving broader evaluation on more diverse domains, model families, and multimodal reasoning settings for future work.
\section{Broader Impacts}
\label{app:border}
\ourmethod aims to improve the reliability and interpretability of CoT reasoning by detecting unfaithful reasoning traces at the instance level. This may benefit applications that require transparent multi-step reasoning, such as education, scientific question answering, and decision-support systems
%  \begin{figure*}[t!]    % 常规操作\begin{figure}开头说明插入图片
% 					% 后面跟着的[htbp]是图片在文档中放置的位置，也称为浮动体的位置，关于这个我们后面的文章会聊聊，现在不管，照写就是了
% 					\centering    
% % 前面说过，图片放置在中间
% 	\includegraphics[width=1.0\columnwidth]{experiments/efficiency_f1_combined_vs_crv.pdf}
		
% 					\caption{}

%       % 整个图片的说明，注释写在{}内
% 					\label{fig:efficiency_runtime_f1}            % 整个图片的标签编号，注意这里跟子图是一样的道理，标签不能重复 
                
% 				\end{figure*}
%%%%%%%%%%%%%%%%%%%%%%%%%%%%%%%%%%%%%%%%%%%%%%%%%%%%%%%%%%%%
% \hfill
% \begin{subfigure}[t]{0.32\textwidth}
%     \centering
%     \includegraphics[width=\linewidth]{experiments/efficiency_f1_combined_vs_crv.pdf}
%     \caption{Placeholder for the third figure.}
%     \label{fig:efficiency_runtime_f1}
% \end{subfigure}
\newpage

\end{document}